\documentclass[10pt,twocolumn,letterpaper]{article}

\usepackage[pagenumbers]{cvpr} 

\usepackage[utf8]{inputenc} 
\usepackage[T1]{fontenc}    
\usepackage{url}            
\usepackage{booktabs}       
\usepackage{amsfonts}       
\usepackage{nicefrac}       
\usepackage{microtype}      
\usepackage{algorithm}
\usepackage{algpseudocode}
\usepackage{amsmath}
\usepackage{amssymb}
\usepackage{mathtools}
\usepackage{subcaption}
\usepackage{graphicx}
\usepackage{graphics}
\usepackage{tikz}
\usepackage{wrapfig}
\usetikzlibrary{backgrounds}
\usepackage{bm}
\usepackage{tcolorbox}
\usepackage{sidecap}
\usepackage{enumitem}
\usepackage{xcolor}
\usepackage{pifont}
\usepackage{setspace}

\definecolor{skyblue}{RGB}{232, 243, 255}

\newcommand{\R}{\mathbb{R}}

\newcommand{\N}{\mathcal{N}}
\newcommand{\I}{\mathbf{I}}

\newcommand{\norm}[1]{\| #1 \|}

\newcommand{\E}{\mathbb{E}}

\newcommand{\m}[1]{\mathbf{#1}}

\newcommand{\methodName}[0]{DiverseFlow}

\definecolor{cvprblue}{rgb}{0.21,0.49,0.74}
\usepackage[pagebackref,breaklinks,colorlinks,allcolors=cvprblue]{hyperref}

\def\paperID{15042} 
\def\confName{CVPR}
\def\confYear{2025}

\title{DiverseFlow: Sample-Efficient Diverse Mode Coverage in Flows}

\author{
Mashrur M. Morshed \hspace{30pt} Vishnu Boddeti\\
Michigan State University \\
{\tt\small \{morshedm, vishnu\}@msu.edu}
}

\begin{document}
\maketitle

\begin{abstract}
Many real-world applications of flow-based generative models desire a diverse set of samples that cover multiple modes of the target distribution. However, the predominant approach for obtaining diverse sets is not sample-efficient, as it involves independently obtaining many samples from the source distribution and mapping them through the flow until the desired mode coverage is achieved. As an alternative to repeated sampling, we introduce \methodName{}: a training-free approach to improve the diversity of flow models. Our key idea is to employ a determinantal point process to induce a coupling between the samples that drives diversity under a fixed sampling budget. In essence, \methodName{} allows exploration of more variations in a learned flow model with fewer samples. We demonstrate the efficacy of our method for tasks where sample-efficient diversity is desirable, such as text-guided image generation with polysemous words, inverse problems like large-hole inpainting, and class-conditional image synthesis.
\end{abstract}
    
\section{Introduction}

Consider the task of text-guided image generation from open-ended prompts, like \textbf{\emph{``A famous boxer''}}.
If we use a generative ordinary differential equation (ODE) to obtain a few images for this given prompt, we may observe something unusual: all the resultant images depict \emph{a dog}, as shown in \Cref{fig:teaser} (top-left).
Such a result is plausible, as the word ``boxer" has multiple meanings: it can either mean a \emph{combat sport athlete} or a particular \emph{dog breed}.
However, for the prompt \textbf{\emph{``A famous boxer''}}, what if the output desired by the user was actually a human athlete, and not a dog?
This situation necessitates obtaining additional samples from the model, until the desired alternate meanings are discovered.
But instead of repeated sampling, can we directly observe more meanings by finding a more \emph{diverse} set?

Beyond the aforementioned example of text-to-image generation from polysemous\footnote{Words or phrases with several meanings.} prompts, sample diversity is a desirable objective for many other tasks that use generative models. These include inverse problems (e.g., large hole filling) and class-conditioned image generation, to name a few. Diversity or mode coverage is a key pillar in the \emph{generative learning trilemma}~\citep{xiao2022tackling}, in addition to fidelity and latency.
For state-of-the-art generative methods such as flow matching models (FM) \citep{lipman2022flow, liu2022flow} and diffusion models (DM) \citep{sohl2015deep,ho2020denoising}, significant work has been done on improving the photorealism of samples and the efficiency of the sampling process \citep{ho2022classifier,karras2022elucidating,lipman2022flow,zheng2023guided,ddim,tong2023improving}. However, relatively little attention has been paid to explicitly enhancing the diversity of generated samples under a limited sampling budget.

\def\rowa{0}
\def\rowb{2.85}
\def\deltarow{4}
\def\cola{0}
\def\deltacol{3.85}

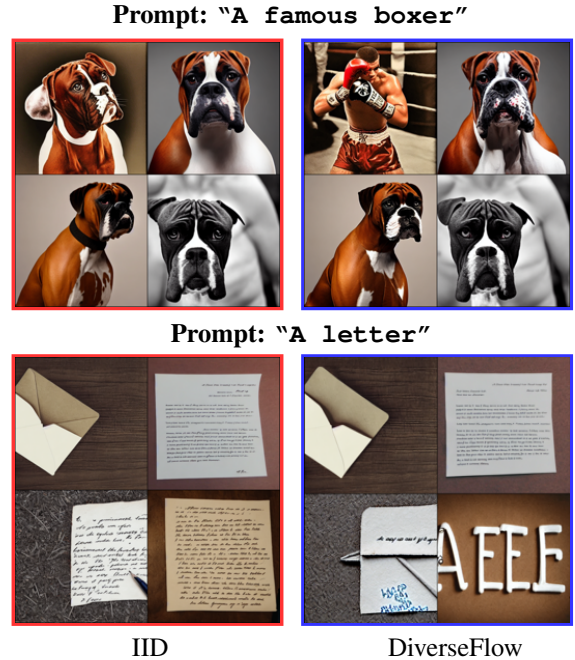
\begin{figure}[t]
\centering

\begin{tikzpicture}
\node[anchor=south west, inner sep=0, line width=0pt] at (\cola + 0.35 * \deltacol, \rowa + 3.35){
   \textbf{Prompt: \texttt{``A famous boxer''}}
};

\node[anchor=south west, draw=red!80, line width=1mm, inner sep=0pt,minimum size=12pt](node:1A) at (\cola, \rowa - 0.4){
    \includegraphics[width=100pt]{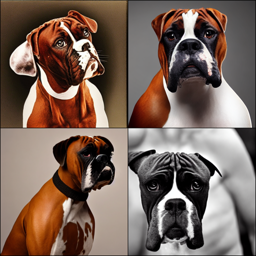}
};

\node[anchor=south west, draw=blue!80, line width=1mm, inner sep=0pt,minimum size=12pt](1B) at (\cola + \deltacol , \rowa - 0.4){
    \includegraphics[width=100pt]{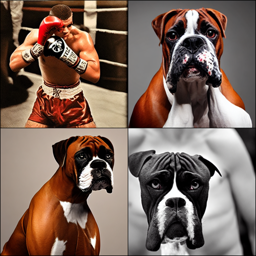}
};

\node[anchor=south west, inner sep=0, line width=0pt] at (\cola + 0.55 * \deltacol, \rowa - 0.4 - 0.5){
   \textbf{Prompt: \texttt{``A letter''}}
};

\node[anchor=south west, draw=red!80, line width=1mm, inner sep=0pt,minimum size=12pt](1C) at (\cola, \rowa - \deltarow - 0.6){
    \includegraphics[width=100pt]{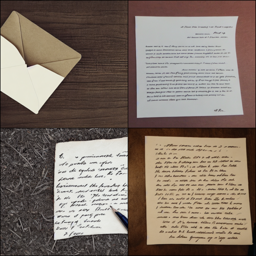}
};
\node[anchor=south west, inner sep=0, line width=0pt] at (\cola + 1.6, \rowa - \deltarow - 1){
   IID
};

\node[anchor=south west, draw=blue!80, line width=1mm, inner sep=0pt,minimum size=12pt](1D) at (\cola + \deltacol, \rowa - \deltarow - 0.6){
    \includegraphics[width=100pt, height=100pt]{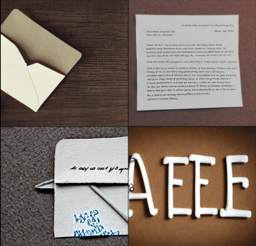}
};

\node[anchor=south west, inner sep=0, line width=0pt] at (\cola + 1.3 * \deltacol, \rowa - \deltarow - 1){
   \methodName{}
};
\end{tikzpicture}
\caption{\small{Text-guided image generation with polysemous words. With the same budget of samples, \methodName{} (right) finds a more diverse set of results compared to IID sampling (left).}}
\label{fig:teaser}
\end{figure}

The standard approach to generate a diverse set of images is to repeatedly obtain independent and identically distributed (IID) samples from a simple or tractable source distribution (e.g., Gaussian distribution), map them to samples in the target distribution, and continue this process until we observe sufficient mode coverage in the target distribution. This process, while effective, is \emph{sample-inefficient}, requiring the generation of more results than necessary. Importantly, the mapping from the source to the target density does not hold a linear relationship; even specifically selecting diverse samples from the source distribution by design does not necessarily yield diverse samples in the target distribution. These limitations naturally raise the following research question.

\begin{tcolorbox}
    \emph{How can we generate diverse samples from the target density under a limited sampling budget?}
\end{tcolorbox}

In this paper, we propose \methodName{}, an inference-time, training-free approach to obtain a diverse set of samples in a desired target density under a fixed sampling budget. We focus on deterministic ODE sampling in continuous-time generative models, specifically FMs, an emerging generative paradigm that enables simulation-free training of continuous normalizing flows (CNFs) and includes diffusion models as a special case.

\methodName{} measures the diversity of a set of samples through the \emph{volume} they span in the target space. A set of similar samples span a lower volume, while a diverse set naturally spans a larger volume. We impose a volume-based gradient constraint on the flow ODE by drawing on determinantal point processes (DPP) \citep{macchi1975coincidence,kulesza2012determinantal}, a probabilistic model arising from quantum physics that exactly describes the Pauli exclusion principle: that no two fermions may occupy the same quantum state.
In \Cref{fig:teaser} consider the images  generated via IID sampling (left column) from a text-to-image generative ODE---for both the prompts \textbf{\emph{``A famous boxer''}} and \textbf{\emph{``A letter''}}, only a single potential meaning is discovered. Unlike the results generated via IID sampling, samples obtained from \methodName{} (right column) span more diverse modes corresponding to the polysemous words in the prompts. For \textbf{\emph{``A famous boxer''}}, the samples show both a dog breed and a human athlete; for \textbf{\emph{``A letter"}}, we observe both \emph{written correspondence} and \emph{alphabet symbols}.

We empirically demonstrate the utility of \methodName{} across several applications where diversity is inherently desirable. First, we use DiverseFlow to perform \textbf{text-guided image synthesis} for words and phrases that may carry a variety of meanings. Second, we perform \textbf{large-hole face inpainting} with occlusion masks covering significant regions of the face that may be important to the person's identity. Third, we apply DiverseFlow on \textbf{class-conditioned image synthesis} and demonstrate that we can more efficiently explore the data space compared to IID sampling. Lastly, to better characterize and explain the behavior of DiverseFlow, we perform several experiments on synthetic 2D densities.

\begin{tcolorbox}[width=\linewidth,
                  colback=skyblue,
                  colframe=skyblue,
                  boxsep=5pt,
                  left=0pt,
                  right=0pt,
                  top=2pt]
\noindent\textbf{Summary of Contributions} 
\begin{enumerate}[leftmargin=0.5cm]
\setlength{\itemsep}{0.0cm}
    \item We present a sample-efficient method to obtain a diverse set of samples from a flow ODE in \Cref{section:method} and demonstrate it qualitatively in \Cref{sec:text-to-image,sec:inpaint,sec:imagenet} and quantitatively in \Cref{sec:flow-types,sec:comp-to-pg}
    \item We provide an empirical analysis that demonstrates that our method is consistent across various flow matching formulations (\Cref{sec:flow-types})
    \item We introduce the task of text-to-image synthesis from prompts with polysemous words in the context of analyzing sample diversity, and provide a comparison with relevant methods (\Cref{sec:comp-to-pg})
    
\end{enumerate}
\end{tcolorbox}
\section{Preliminaries}
\subsection{Flow Matching}

Many generative models can be considered as a \emph{transport map} from some easy-to-sample source distribution to an empirically observed yet unknown target distribution. Recent successes in generative modeling represent this transport map in the form of continuous-time processes, such as stochastic differential equations (SDEs) \citep{song2020score,ho2020denoising}, or ordinary differential equations (ODEs) \citep{lipman2022flow,liu2022flow,albergo2023stochastic}.
Although diffusion models are formulated as SDEs, a significant body of research focuses on converting the diffusion SDE to a deterministic ODE at inference time for faster inference.
The diffusion ODE, or probability flow ODE, is a particular case of continuous normalizing flows (CNFs).
Flow Matching (FM) \citep{lipman2022flow,liu2022flow,albergo2023stochastic} is motivated by the idea of directly training CNFs in a scalable and simulation-free manner, just like diffusion models.
Moreover, many recent text-to-image generative models, such as Stable Diffusion 3 \citep{esser2024scaling}, adopt the FM framework.
As such, we present our approach primarily in the context of FM, and our findings can be extended to diffusion and score-based generative models in a straightforward manner.

A CNF reshapes a prior source density $p_0$ to the empirically observed target density $p_1$ with an ODE of the form:
\begin{equation}
    d\m{x}_t = v_{\theta}(\m{x}_t, t)dt, \quad \m{x}_0 \sim p_0
    \label{eqn:flow-ode}
\end{equation}
\noindent where $v_{\theta}$ is a time-dependent velocity field whose parameters $\theta$ are learned; we interchangeably use the notation $v_t$ to imply $v_{\theta}(\cdot, t)$. It becomes possible to obtain samples from $p_1$ by integrating \Cref{eqn:flow-ode} over time, i.e. by starting at $\m{x}_0 \sim p_0$ for $t=0$ and solving the ODE till $t=1$. As our approach is training-free, we do not elaborate on the details of learning to regress the vector field $v_t$; we encourage interested readers to refer to the works of \citet{lipman2022flow} and \citet{tong2023improving} for a primer on training FMs.

At any timestep $t$ during sampling, an intermediate sample $\m{x}_t$ in the flow trajectory can be used to obtain an approximation of the target as follows:
\begin{equation}
    \hat{\m{x}}_1 = \m{x}_t + v_{\theta}(\m{x}_t, t) (1 - t)
    \label{eqn:pred-x1}
\end{equation}
\Cref{eqn:pred-x1} is equivalent to simply taking a large Euler step at any time instance $t$ and is naturally more accurate as $t$ approaches $t=1$. Further, \Cref{eqn:pred-x1} is also well suited for ODEs with `straight' paths, where the direction of the time-varying velocity $v_t$ remains near-constant in time (such as the work of \citet{liu2022flow}). Similarly, we can estimate the source sample by simply taking a step in the reverse direction:
\begin{equation}
    \hat{\m{x}}_0 = \m{x}_t - v_{\theta}(\m{x}_t, t) t
    \label{eqn:pred-x0}
\end{equation}

\subsection{Determinantal Point Processes}
Determinantal point processes (DPPs) \citep{macchi1975coincidence,borodin2000distributions,kulesza2012determinantal} are probabilistic models of repulsion between points. They were originally termed as \textit{fermion processes} as they describe the Pauli exclusion principle or antibunching effect in fermions. To define a DPP, we must first consider a set of points, $\mathcal{Y}$, and a point process $\mathcal{P}(\mathcal{Y})$---a probability measure on $2^\mathcal{Y}$ (the set of all possible subsets of $\mathcal{Y})$. $\mathcal{P}$ is \textit{determinantal} when the probability of choosing a random subset $Y \subset \mathcal{Y}$ according to $\mathcal{P}$ is given by:
\begin{align}
    \mathcal{P}(Y \subset \mathcal{Y}) = \frac{\det(\m{L}_Y)}{\sum_{Y \subset \mathcal{Y}} \det(\m{L}_Y)} = \frac{\det(\m{L}_Y)}{\det(\m{L} + \I)}
    \label{eqn:background-dpp}
\end{align}
\noindent where $\m{L} \in \R^{\mid \mathcal{Y} \mid \times \mid \mathcal{Y} \mid}$ is a kernel matrix, and $\m{L}_Y$ is the sub-kernel matrix indexed by the elements of $Y$. \Cref{eqn:background-dpp} has an intuitive geometric interpretation if we consider the kernel $\m{L}$ to be constructed from cosine similarity: the determinant of $\m{L}_Y$ is the Gram-determinant, describing the squared volume of the $N$-dimensional parallelotope spanned by the set of vectors $Y$. Thus, a DPP naturally assigns higher probabilities to more orthogonal (and thus diverse) subsets that span larger volumes. We leverage DPPs to define a coupled likelihood measure over a set of samples in a flow trajectory.
\section{Related Work}

Efficiently finding diverse sets is useful in several application areas of machine learning. For instance, \citet{batra2012diverse} show that the M-Best MAP (maximum a posteriori) solutions in Markov random fields are often distant from the ground truth and highly similar. They thus propose the \textit{Diverse M-Best} problem---finding a set of $M$ highly probable solutions satisfying some minimum dissimilarity threshold---that partly inspires our study in \Cref{section:problem-setting}. \cite{yuan2019diverse} utilize DPPs in conjunction with variational autoencoders (VAE) for diverse trajectory forecasting; a set of diverse future pedestrian trajectories improves safety-critical perception systems in autonomous vehicles. Motivated by potential drug discovery and material design applications, \citet{jain2023multi} propose finding diverse Pareto-optimal candidates in a multi-objective setting with generative flow networks.

The work by \citet{corso2023particle} which explores diverse non-IID sampling for diffusion models is most similar in spirit to \methodName{}. However, \methodName{} is notably different in the following aspects:
(1) Our diversity objective is derived from determinantal point processes, a diversity-promoting probability measure of the joint occurrence of a set of samples. \citet{corso2023particle} is instead inspired by stein variational gradient descent (SVGD) \citep{liu2016stein}.
2) The diversity measure in \methodName{} (volume, or determinant of similarity kernel) assigns a zero likelihood to a set if any duplicate elements are present; presence of duplicates is tolerated in the diversity measure in Particle Guidance  (row-wise sum of similarity kernel)
\section{Diverse Source Samples Do Not Yield Diverse Target Samples}
\label{section:problem-setting}

\setlength{\belowcaptionskip}{0ex}
\begin{figure}[!h]
    \centering
    \begin{subfigure}[t]{0.19\textwidth}
        \centering
        \includegraphics[width=\textwidth, height=\textwidth]{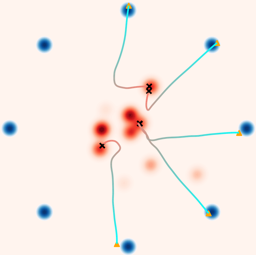}
        \caption{Conditional Flow Matching (CFM) \citep{lipman2022flow}}
        \label{subfig:cfm}
    \end{subfigure}
    \begin{subfigure}[t]{0.19\textwidth}
        \centering
        \includegraphics[width=\textwidth,  height=\textwidth]{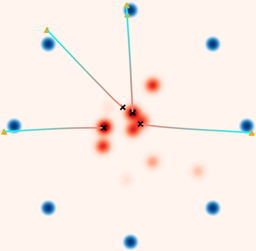}
        \caption{Mini-batch Optimal Transport (MB-OT) \citep{tong2023improving}}
        \label{fig:subfigc}
    \end{subfigure}
    
    \begin{subfigure}[t]{0.19\textwidth}
        \centering
        \includegraphics[width=\textwidth,  height=\textwidth]{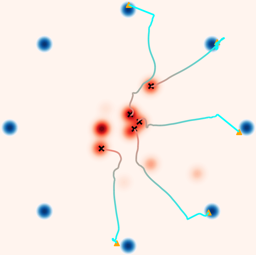}
        \caption{\methodName{} on CFM}
        \label{fig:subfigd}
    \end{subfigure}
    \begin{subfigure}[t]{0.19\textwidth}
        \centering
        \includegraphics[width=\textwidth,  height=\textwidth]{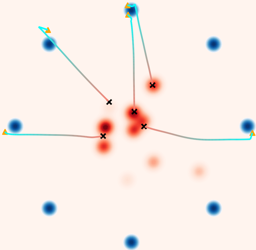}
        \caption{\methodName{} on MB-OT}
        \label{fig:subfige}
    \end{subfigure}
    \caption{Finding $K=5$ samples from the target distribution with $N=10$ modes. In the example, \textbf{only 3 modes} are found by 5 IID samples (a, b). \methodName{} discovers \textbf{5 modes} with the same sampling budget in (d, e).}
    \label{fig:mainfig}
\end{figure}

\textbf{Problem Setting:} \quad We start with a synthetic example to illustrate our problem of interest. Consider that we have empirical observations from a target distribution $\pi_1 \in \R^2$, which is a random mixture of Gaussians, such as the example shown in \Cref{fig:modefinding}. We design $\pi_1=\sum_{i=1}^N w_i\mathcal{N}(\bm{\mu}_i,\sigma^2_i\bm{I})$ to contain $N=10$ randomly selected modes $\mathcal{N}(\bm{\mu}_i,\sigma^2_i\bm{I})$, each with a random mixture weight $w_i$; we observe that in our example, there are 6 high probability modes and 4 low probability ones. Suppose we have a sampling budget of $K$ samples. This leads to three possible scenarios: (i) $K < N$, (ii) $K = N$, and (iii) $K > N$. Among the aforementioned, case (i) (fewer samples than modes) is the most likely characteristic of any real-world dataset.

Let us have a prior distribution $\pi_0$ and some generative model $\Psi$, such that, in the limit of infinite samples, $\Psi(x_0 \sim \pi_0) \sim \pi_1$. Then, the objective of \emph{sample-efficient diverse sampling} is to obtain samples from $\min(K, N)$ modes from $\pi_1$, given a fixed set of $K$ samples in $\pi_0$.

If diverse samples are desired from the target density of the flow, one may make the elementary assumption that \emph{if the particles are distant at the source distribution, after being transported by the flow, they remain distant in the target distribution}. This assumption is not necessarily true, as we show in \Cref{fig:mainfig}. By design, we choose a uniform mixture of eight Gaussians as the source $\pi_0$ to obtain diverse source samples. In \Cref{subfig:cfm}, we can observe that source points from distinct modes can still converge to the same target mode with IID sampling. Thus, an alternative procedure is necessary to obtain a diverse set from a flow in a sample-efficient manner. We further explore this toy problem in \Cref{sec:flow-types}.
\section{Diverse Sampling from Flows\label{section:method}}
From \Cref{fig:mainfig}, we observe that independently (or heuristically) chosen source samples may not map to a diverse set of target samples. In this case, we can select a new set of source samples and repeat the sampling process till eventually covering at least $K$ modes. However, this approach does not satisfy our fixed sampling budget constraint. An alternative solution to repeated independent sampling is defining and leveraging a diversity measure of the target samples to drive sample diversity. For the set of source samples $\{\m{x}_0^{(1)}, \m{x}_0^{(2)}, \dots, \m{x}_0^{(k)}\}$, we could optimize a set of perturbations $\{\bm{\delta}^{(1)},\bm{\delta}^{(2)}, \dots, \bm{\delta}^{(k)}\}$ such that the new set $\{\m{x}_0^{(1)} + \bm{\delta}^{(1)}, \m{x}_0^{(2)} + \bm{\delta}^{(2)}, \dots, \m{x}_0^{(k)} + \bm{\delta}^{(k)}\}$ maps to a diverse set of target particles. However, this approach would require multiple simulations of the whole ODE and backpropagating over all the timesteps, which increases the computational complexity of the sampling process over the standard IID sampling.

This leads us to our proposed approach: we avoid multiple simulations and instead optimize the flow trajectory for diversity \emph{while solving the ODE}.
For any sample in the flow trajectory $\m{x}_t$, we have an estimate of the target sample $\hat{\m{x}}_1$ through \Cref{eqn:pred-x1}.
Suppose we have a differentiable objective $\mathcal{L}(\{\hat{\m{x}}_1^{(1)}, \hat{\m{x}}_1^{(2)}, \dots, \hat{\m{x}}_1^{(k)}\})$ that assigns a likelihood to the joint outcome $\{\hat{\m{x}}_1^{(1)}, \hat{\m{x}}_1^{(2)}, \dots, \hat{\m{x}}_1^{(k)}\}$. Further, let $\mathcal{L}$ assign a higher likelihood if the joint outcome is a diverse set, and a diminished likelihood if the set is similar.  We can then leverage $\mathcal{L}$ to drive diversity among the target samples by modifying the flow velocity of the $i$-th particle as,
\begin{equation}
    \tilde{v}_t^{(i)} = v_t^{(i)} - \gamma(t) \nabla_{\m{x}_t^{(i)}} \log \mathcal{L}(\{\hat{\m{x}}_1^{(1)}, \hat{\m{x}}_1^{(2)}, \dots, \hat{\m{x}}_1^{(k)}\})
    \label{eqn:modified-flow}
\end{equation}
\noindent where $\gamma(t)$ is a time-varying scale that controls the strength of the diversity gradient. Setting $\gamma(t)=0$ reduces to the standard IID sampling scenario, while $\gamma(t) > 0$ will encourage diversity between the generated samples.
In practice, $\gamma(t)$ follows the schedule of the probability path normalized by the norm of the DPP gradient.

\subsection{Determinantal Gradient Constraints}

We desire objective $\mathcal{L}$ in \Cref{eqn:modified-flow} to be higher if the items in the set are diverse and lower if they are similar to each other. We interpret diversity in terms of the \emph{volume} spanned by the set. Consider that we have $k$ samples in $\R^d$ (assume $k < d$). An objective that prefers diversity can be defined as the volume of the $k$-dimensional parallelotope in $\R^d$ spanned by the sample vectors; this volume becomes diminished when there are similar samples (and even zero, for identical samples). The determinant describes volumes well; a diverse set must span a large volume in the sample space and have a corresponding large determinant.

To define a measure over a set of samples, we draw on the idea of determinantal point processes (DPP). We first define a kernel $\m{L}(\{\hat{\m{x}}_1^{(1)}, \hat{\m{x}}_1^{(2)}, \dots, \hat{\m{x}}_1^{(k)}\})$ as follows:
\begin{equation}
    \m{L}^{(ij)} = \exp \left(-h \frac{\|\hat{\m{x}}_1^{(i)} - \hat{\m{x}}_1^{(j)}\|^2_2}{\text{med}(\m{U}(\m{D}))}\right)
    \label{eqn:dpp-kernel}
\end{equation}
\noindent where $\m{D}$ denotes a distance matrix with $\m{D}_{ij} = \|\m{x}^{(i)} - \m{x}^{(j)}\|^2_2$, $\m{U}(\m{D})$ denotes the upper triangle entries of $\m{D}$, $h$ denotes a kernel spread parameter, and $\text{med}(\m{U}(\m{D}))$ denotes the median of those entries. Given $\m{L}$, we may define a DPP-based likelihood as:

\begin{equation}
\begin{split}
    \mathcal{L}(\{\hat{\m{x}}_1^{(1)}, \hat{\m{x}}_1^{(2)}, \dots, \hat{\m{x}}_1^{(k)}\}) &= \frac{\det(\m{L})}{\det(\m{L} + \I)} \\
    &= \prod_{a=1}^{k} \frac{\lambda(\m{L})_a}{1 + \lambda(\m{L})_a}
    \label{eqn:dpplikelihood}
\end{split}
\end{equation} 

\noindent where $\lambda(\m{L})_a$ is the $a^{\text{th}}$ eigenvalue of the kernel $\m{L}$. The log-likelihood is then,
\begin{equation}
    \mathcal{LL} = \log\mathcal{L} = \log\det(\m{L}) - \log\det(\m{L} + \I)
    \label{eqn:dppnll}
\end{equation}

Note that the Euclidean distance $\|\hat{\m{x}}_1^{(i)} - \hat{\m{x}}_1^{(j)}\|^2_2$ is not very meaningful in the high-dimensional raw image space \citep{aggarwal2001surprising}. Therefore, in practice, the distance should be computed in a robust feature space, i.e., $\|F(\hat{\m{x}}_1^{(i)}) - F(\hat{\m{x}}_1^{(j)})\|^2_2$, where $F$ is some domain-specific feature extractor, such as the vision transformer (ViT)~\citep{dosovitskiy2020image} for images.

\textbf{Quality Constraint:}
The DPP defined in \Cref{eqn:dpplikelihood} acts as a repulsive force that is unaware of sample quality. A quality term can be incorporated into the DPP kernel to regularize the trajectory diversification. Although flows can be defined between any arbitrary two distributions, let us consider the special case when the source is a Gaussian, i.e., $p_0 \sim \N(0, \I)$. Suppose we have a quality vector $\m{q}_t = \{q^{(1)}(t), {q}^{(2)}(t), \dots, {q}^{(k)}(t)\}$, where any $q^{(i)}(t) \in [0, 1]$. We can then define a new kernel $\m{L}_q = \m{L} \odot \m{q}_t\m{q}_t^T$, where each $q^{(i)}(t)$ penalizes a sample $\m{x}_t^{(i)}$ if it deviates too much from the flow. To define this, we obtain an estimate of the source sample $\hat{\m{x}}_0^{(i)}(t)$ for any given sample $\m{x}_t^{(i)}$ via \Cref{eqn:pred-x0}, and check if it lies within a desired percentile-radius $\rho$ of the Gaussian $p_0$. Specifically, we define the time-dependent sample quality as
\begin{equation}
    q^{(i)}(t) = \begin{cases} 1 &  \text{if } \|\hat{\m{x}}_0^{(i)}(t)\|_2^2 \leq \rho^2\\
    \max\left(\epsilon, e^{-\left(\|\hat{\m{x}}_0^{(i)}(t)\|_2^2 - \rho^2\right)} \right) & \text{otherwise}
    \end{cases}
    \label{eqn:quality}
\end{equation}
\noindent where $\epsilon$ is a `minimum quality' we assign to prevent a zero determinant.

\subsection{Coupled Ordinary Differential Equations}

At any timestep $t$, the measure of diversity in \Cref{eqn:dppnll} can be adopted to modify the flow of the $i$-th particle. We compute the gradient of the samples with respect to the diversity measure and use it to modify the ODE as follows:
\begin{equation}
    d\m{x}_t^{(i)} = \left[v_{\theta}(\m{x}_t^{(i)},t) - \gamma(t) \nabla_{\m{x}_t^{(i)}} \log \mathcal{L}(\{\hat{\m{x}}_1^{(1)}, \dots, \hat{\m{x}}_1^{(k)}\})\right]dt
    \label{eqn:guidance}
\end{equation}
Where $\gamma(t)$ is a time-varying scaling factor. Unlike the IID sampling scenario where we have $K$ independent ODEs, \Cref{eqn:guidance} corresponds to a system of coupled non-linear ordinary differential ordinary equations. To see this, first note that the estimate $\hat{\m{x}}_1^{(i)}$ depends on the current sample $\m{x}_t^{(i)}$ i.e., $\hat{\m{x}}_1^{(i)} = \m{x}_t^{(i)} + v_{\theta}(\m{x}_t^{(i)},t)(1-t)$. Second, the DPP log-likelihood $\mathcal{LL}(\{\hat{\m{x}}_1^{(1)}, \hat{\m{x}}_1^{(2)}, \dots, \hat{\m{x}}_1^{(k)}\})$ induces a time-dependent coupling between the $K$ trajectories of $\m{x}^{(i)}_t, i=1,\dots, K$ and seeks to diversify the target samples. Although higher-order ODE solvers~\citep{karras2022elucidating} can be employed to solve the coupled ODEs, we use the standard Euler method.
\section{Experiments\label{sec:experiments}}

We demonstrate the utility of \methodName{} in flow-based generative models by considering three applications where sample diversity is naturally desirable: text-guided image generation with polysemous words, large-hole inpainting, and class-conditional image generation. We also analyze the effect of \methodName{} on different flow matching formulations w.r.t. its ability to span diverse modes through a synthetically constructed 2D density example.

\def\rowa{0}
\def\deltarow{3.7}
\def\cola{0}
\def\deltacol{3.6}

\begin{figure}

\centering
\captionsetup{type=figure}
\begin{tikzpicture}

\node[anchor=south west,  inner sep=0, draw=red!80, line width=1mm ,minimum size=12pt,](1A) at (\cola, \rowa){
    \includegraphics[width=100pt]{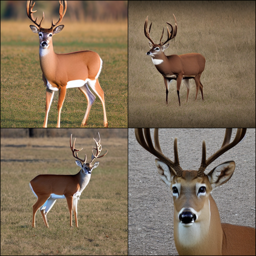}
};
\node[anchor=south west, inner sep=0, draw=blue!80, line width=1mm ,minimum size=12pt,](1B) at (\cola + \deltacol, \rowa){
    \includegraphics[width=100pt]{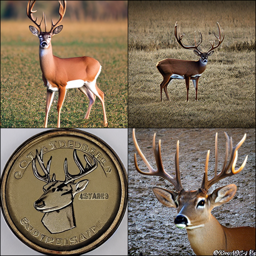}
};
\node[anchor=south west, inner sep=0, line width=0pt](1A) at (\cola + 2.4, \rowa - 0.5){
 (a) \texttt{``A buck''}
};

\node[anchor=south west,  inner sep=0, draw=red!80, line width=1mm ,minimum size=12pt,](1A) at (\cola, \rowa - \deltarow - 0.5){
    \includegraphics[width=100pt]{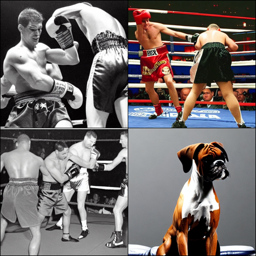}
};
\node[anchor=south west, inner sep=0, draw=blue!80, line width=1mm ,minimum size=12pt,](1B) at (\cola + \deltacol, \rowa - \deltarow - 0.5){
    \includegraphics[width=100pt]{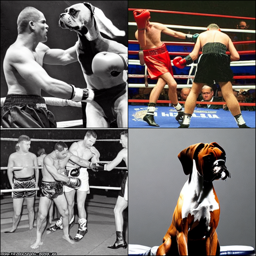}
};
\node[anchor=south west, inner sep=0, line width=0pt](1A) at (\cola + 1.6, \rowa - \deltarow - 1){
 (b) \texttt{``A famous boxer''}
};

\node[anchor=south west,  inner sep=0, draw=red!80, line width=1mm ,minimum size=12pt,](1A) at (\cola, \rowa - 2 * \deltarow - 1){
    \includegraphics[width=100pt]{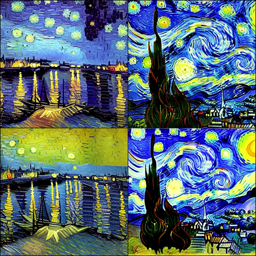}
};
\node[anchor=south west,  inner sep=0, draw=blue!80, line width=1mm ,minimum size=12pt,](1A) at (\cola + \deltacol, \rowa - 2 * \deltarow - 1){
    \includegraphics[width=100pt]{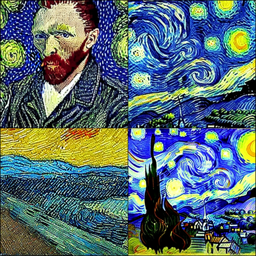}
};
\node[anchor=south west, inner sep=0, line width=0pt](1A) at (\cola + 1.4,\rowa - 2 * \deltarow - 1.5){
 (c) \texttt{``Van Gogh painting''}
};

\end{tikzpicture}
\captionof{figure}{For each prompt, the left image ({\color{red!80} red box}) denotes standard IID sampling with classifier-free guidance, while the right image ({\color{blue!80} blue box}) shows the result after incorporating \methodName{}. \methodName{} finds more diverse sets given the same source points---clearly distinguishable as new semantic meanings in the case of prompts with multiple meanings.}
\label{fig:polysemous}
\end{figure}

\subsection{Image Generation from Polysemous Prompts\label{sec:text-to-image}}

In text-to-image generation, the conditional data distribution corresponding to a text prompt may contain many variations, and it is a desirable objective to generate images that span those variations in a sample-efficient manner. As we have previously shown in \Cref{fig:teaser}, the result desired by a user (a human boxer athlete) may not be the result generated by the model (a dog breed); if diverse results are obtained, it allows a user to obtain the desired results with fewer attempts.

We pose a scenario where diverse sets are easily observable: when an open-ended text prompt is \emph{polysemous} and carries \emph{multiple meanings}, such as the examples we show in \Cref{fig:teaser} and \Cref{fig:polysemous}. In \Cref{fig:polysemous}(a), the prompt ``A buck'' may commonly refer to a male deer. However, it may also informally refer to a United States dollar. Using the \emph{same} four source points, which are deterministically mapped to four deer images by IID sampling, \methodName{} finds a different set of samples---one that includes a dollar-like coin, albeit embossed with a deer head. We also observe minor differences between the two sets of images, such as changes in pose and background in the top-right and bottom-right deers. For \Cref{fig:polysemous}(c), while `Van Gogh painting' is not quite a polysemous word, it can still have two interpretations: a painting \emph{painted by} Van Gogh, or a painting \emph{of} Van Gogh. The regular samples contain minimal diversity, as they include two sets of repeated paintings of Van Gogh. With \methodName{}, not only can we get a set of four distinct paintings, but we also have a portrait of Van Gogh, which is one of the additional meanings of the prompt. We present additional qualitative samples in the supplementary material.

\begin{figure*}[ht]
    \centering
    \includegraphics[width=\textwidth]{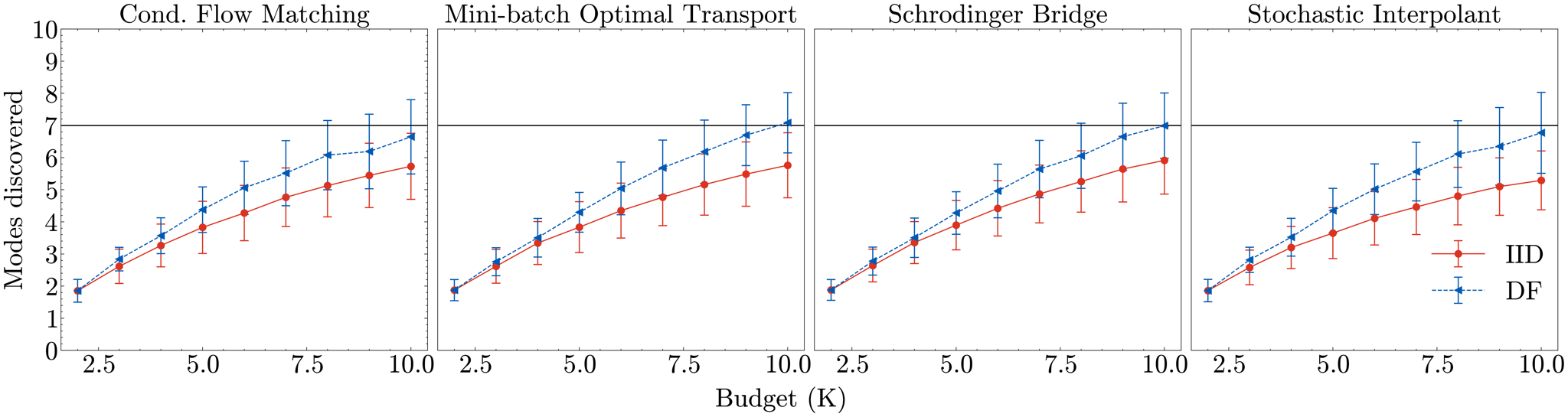}
    \caption{Comparing different FM formulations in terms of the number of modes spanned by IID sampling versus with DiverseFlow. More details about the experiment are provided in the supplementary.}
    \label{fig:modefinding}
\end{figure*}

\subsection{Diverse Inpainting on Faces}
\label{sec:inpaint}

Another inverse problem where diverse solutions are desirable is face inpainting, where we seek to inpaint the missing parts of the face with diverse plausible facial textures and structures. To demonstrate inpainting with FM models, we first incorporate Manifold Constrained Gradient (MCG) \citep{chung2022improving} in an off-the-shelf unconditional Rectified-Flow model. In addition to the manifold constraints, we employ determinantal gradient constraints to enhance diversity (further detailed in \Cref{algo:inpainting}). In \Cref{fig:inpaint} (b), we observe that the inpainted faces of the four women have similar expressions (largely neutral). \methodName{} improves the diversity of the set by yielding a highly different expression in the top-right image. In (d) and (e), we also observe changes in facial hair and expressions due to diversification.  

\def\rowa{0}
\def\deltarow{2.9}
\def\cola{0}
\def\deltacol{2.8}

\def\W{75pt}
\def\H{75pt}

\begin{center}
\centering
\captionsetup{type=figure}
\begin{tikzpicture}

\node[anchor=south west, dashed, inner sep=0, draw=black!80, line width=0.5mm ,minimum size=12pt,](1A) at (\cola-0.5*\deltacol, \rowa + 0.475*\deltarow){
    \includegraphics[width=\W/2, height=\H/2]{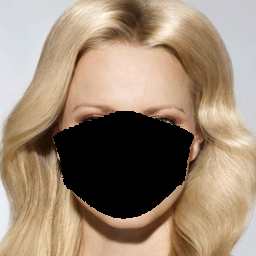}
};

\node[anchor=south west, dashed, inner sep=0, draw=black!80, line width=0.5mm ,minimum size=12pt,](1A) at (\cola-0.5*\deltacol, \rowa + 1.475*\deltarow){
    \includegraphics[width=\W/2, height=\H/2]{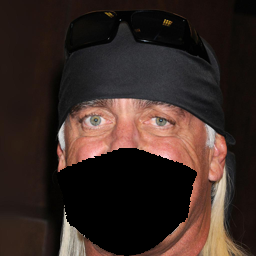}
};


\node[anchor=south west,  inner sep=0, draw=red!80, line width=1mm ,minimum size=12pt,](1A) at (\cola, \rowa){
    \includegraphics[width=\W, height=\H]{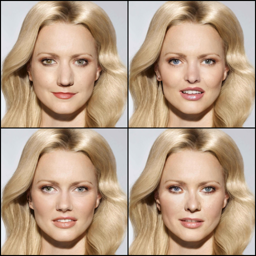}
};
\node[anchor=south west,  inner sep=0, draw=blue!80, line width=1mm ,minimum size=12pt,](1B) at (\cola + \deltacol, \rowa){
    \includegraphics[width=\W, height=\H]{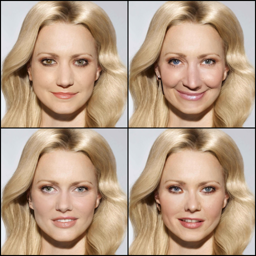}
};

\node[anchor=south west,  inner sep=0, draw=red!80, line width=1mm ,minimum size=12pt,](1A) at (\cola * \deltacol, \rowa + \deltarow){
    \includegraphics[width=\W, height=\H]{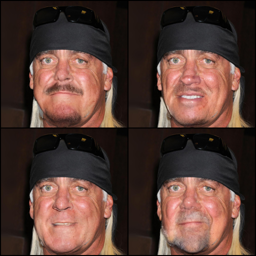}
};

\node[anchor=south west,  inner sep=0, draw=blue!80, line width=1mm ,minimum size=12pt,](1B) at (\cola + \deltacol, \rowa + \deltarow){
    \includegraphics[width=\W, height=\H]{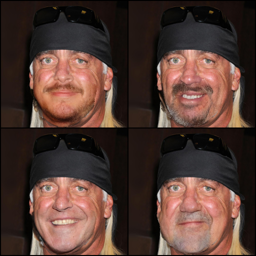}
};


\node[anchor=south west, dashed, inner sep=0, draw=green!80, line width=0.5mm ,minimum size=12pt,](1A) at (\cola-0.5*\deltacol, \rowa){
    \includegraphics[width=\W/2, height=\H/2]{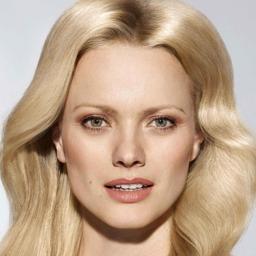}
};

\node[anchor=south west, dashed, inner sep=0, draw=green!80, line width=0.5mm ,minimum size=12pt,](1A) at (\cola-0.5*\deltacol, \rowa + \deltarow){
    \includegraphics[width=\W/2, height=\H/2]{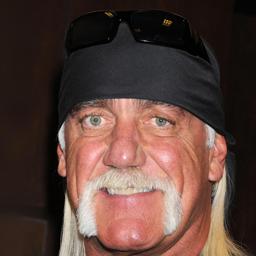}
};

\node[anchor=south west, inner sep=0, line width=0pt](1A) at (\cola - 0.8, \rowa - 0.5){
    (a)
};
\node[anchor=south west, inner sep=0, line width=0pt](1A) at (\cola + 1.1, \rowa - 0.5){
    (b)
};
\node[anchor=south west, inner sep=0, line width=0pt](1A) at (\cola + 1.1 + \deltacol, \rowa - 0.5){
    (c)
};

\end{tikzpicture}
\captionof{figure}{\footnotesize Inpainting on CelebAHQ-$256\times256$. (a) Dashed boxes show masked input (top) and ground truth (bottom) respectively. (b) RectifiedFlow \citep{liu2022flow} + MCG \citep{chung2022improving} (c) With \methodName{}.}

\label{fig:inpaint}

\end{center}

\def\rowa{0}
\def\deltarow{2.75}
\def\cola{0}
\def\deltacol{2.75}

\begin{center}
\centering
\captionsetup{type=figure}
\begin{tikzpicture}
\node[anchor=south west,  inner sep=0, draw=yellow!80, line width=0.5mm ,minimum size=12pt,](1A) at (\cola, \rowa){
    \includegraphics[width=75pt,height=75pt]{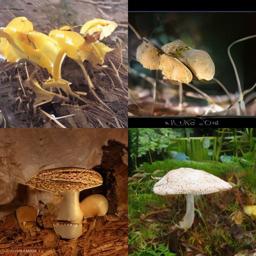}
};
\node[anchor=south west, inner sep=0, draw=red!80, line width=0.5mm ,minimum size=12pt,](2A) at (\cola+\deltacol, \rowa){
    \includegraphics[width=75pt,height=75pt]{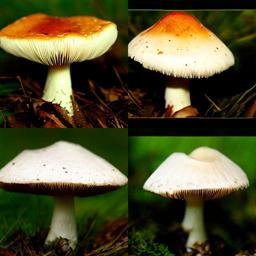}
};

\node[anchor=south west, inner sep=0, draw=blue!80, line width=0.5mm ,minimum size=12pt,](3A) at (\cola+2*\deltacol, \rowa){
    \includegraphics[width=75pt]{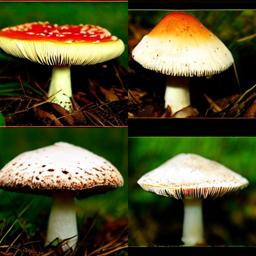}
};

\node[anchor=south west, inner sep=0, draw=yellow!80, line width=0.5mm ,minimum size=12pt,](1A) at (\cola, \rowa + \deltarow ){
    \includegraphics[width=75pt,height=75pt]{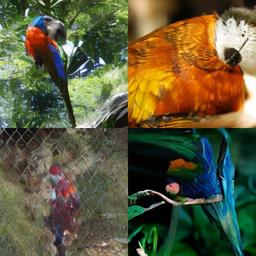}
};

\node[anchor=south west, inner sep=0, draw=red!80, line width=0.5mm ,minimum size=12pt,](1A) at (\cola + \deltacol, \rowa + \deltarow ){
    \includegraphics[width=75pt,height=75pt]{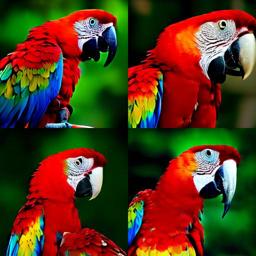}
};

\node[anchor=south west, inner sep=0, draw=blue!80, line width=0.5mm ,minimum size=12pt,](1A) at (\cola + 2 * \deltacol, \rowa + \deltarow ){
    \includegraphics[width=75pt,height=75pt]{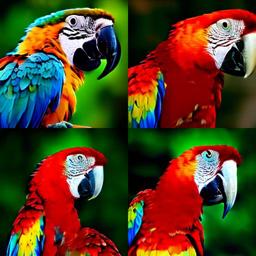}
};

\node[anchor=south west, inner sep=0, line width=0pt,](1A) at (\cola + 0.3 * \deltacol, \rowa-0.5){
    No CFG
};
\node[anchor=south west, inner sep=0, line width=0pt,](1A) at (\cola + 1.2 * \deltacol, \rowa-0.5){
    With CFG
};
\node[anchor=south west, inner sep=0, line width=0pt,](1A) at (\cola + 2.2 * \deltacol, \rowa-0.5){
    \methodName{}
};

\end{tikzpicture}
\captionof{figure}{\footnotesize{Class-conditional ImageNet samples from LFM \citep{dao2023flow}. We show samples for two classes, (top row) `Mushroom' (class 947) and (bottom row) `Macaw' (class 88). (a, d) No CFG. (b, e) LFM with CFG. (d, f) LFM with CFG and \methodName{}.}}
\label{fig:imagenet}

\end{center}

\begin{figure}[!ht]
\begin{minipage}[b]{.46\linewidth}
    \centering
    \includegraphics[width=\linewidth, height=0.85\linewidth]{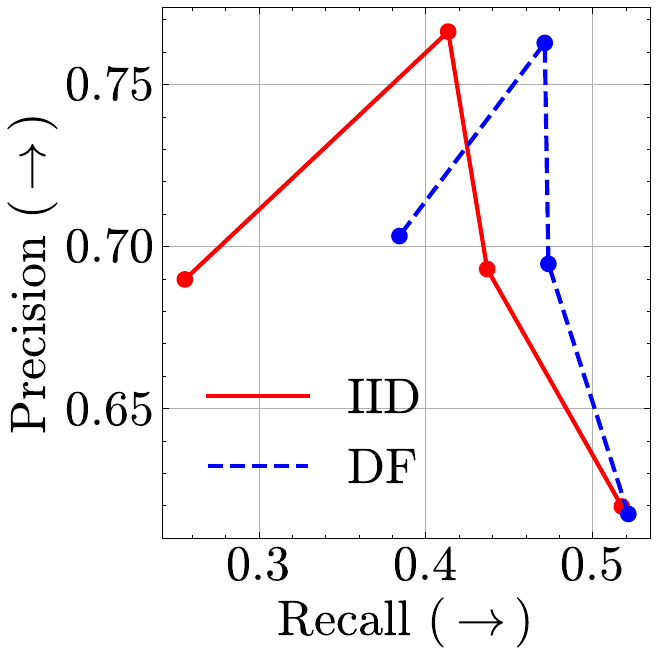}
    \caption{\footnotesize Precision vs. recall on ImageNet-256.}
    \label{fig:prc}
\end{minipage}\hfill
\begin{minipage}[b]{.49\linewidth}
\centering
\tabcolsep=0.11cm
\resizebox{\columnwidth}{!}{
\begin{tabular}{|c|c|c|c|}

\hline
\textbf{Method} &  \textbf{CFG} & \textbf{P ($\uparrow$)} & \textbf{R ($\uparrow$)} \\
\hline
LFM & 1 & 0.62                 & 0.52                        \\
LFM & 1.5 & 0.69                 & 0.44                        \\
LFM & 2 & 0.77                 & 0.41                        \\
LFM & 4 & 0.69                 & 0.26                        \\
\hline
LFM + DF & 1 & 0.62                   & 0.52                       \\
LFM+DF & 1.5 & 0.69                   & 0.47                       \\
LFM+DF & 2 & 0.76                   & 0.46                       \\
LFM+DF & 4 & 0.70                   & 0.38                       \\
\hline

\end{tabular}
}
\captionof{table}{\footnotesize Precision (P) and recall (R) score of \methodName{} and baselines.}
\label{table:precisionrecall}
\end{minipage}
\end{figure}

\subsection{Diverse Class-Conditional Image Synthesis\label{sec:imagenet}}

Suppose we can access a class-conditioned flow matching (FM) model trained on an unknown image dataset. To explore the \emph{unobservable} true dataset, we may use a set of class-conditional samples from the FM model. We adopt a latent flow matching (LFM) model \citep{dao2023flow}, trained to generate $256 \times 256$ resolution images from the ImageNet \citep{deng2009imagenet} dataset. Much like latent diffusion, LFM employs classifier-free guidance to create high-quality samples. However, this naturally poses a cost to diversity, as we show in \Cref{fig:imagenet}.

By incorporating \methodName{}, we can maintain the high quality of the samples and simultaneously explore more modes in the dataset. In \Cref{fig:imagenet}, we demonstrate two ImageNet classes that may have diversity: `Mushroom' and `Macaw.' For mushrooms, we observe that LFM primarily generates two species of mushrooms. However, by applying \methodName{}, we successfully find a new species within our limited set: an \emph{Amanita muscaria}, also known as the \emph{fly agaric}---easily distinguishable by the white spots on its red cap. In another example, we see that while LFM generates four scarlet macaws, using the same source samples, \methodName{} helps us find a different blue and yellow macaw. We further evaluate the precision and recall \citep{kynkaanniemi2019improved} for ImageNet-256 synthesis in \Cref{fig:prc}, showing that \methodName{} improves the recall while retaining similar precision, suggesting improved diversity.

\subsection{\methodName{} Across Various FM Formulations}
\label{sec:flow-types}

We ask the question: does \methodName{} remain consistent across different formulations of flow matching?
We adopt the same toy bivariate mixture of Gaussian distributions as in the example shown in \Cref{fig:mainfig}, across four different FM formulations: (i) Conditional Flow Matching (CFM) \citep{lipman2022flow}, (ii) Mini-batch Optimal Transport CFM (MB-OT) \citep{tong2023improving}, (iii) Schr\"{o}dinger Bridge CFM (SB-CFM) \citep{tong2023simulation}, and (iv) Stochastic Interpolants (SI-CFM) \citep{albergo2023stochastic}. We then perform a numerical experiment to quantify the average number of modes discovered by each FM variant with increasing the sampling budget $K$. \Cref{fig:modefinding} reports the results. For a maximum sampling budget of $10$, MB-OT discovers only $5.64$ modes on average, which is expected since the dataset contains 6 high-probability modes. By incorporating \methodName{}, we can find $7.11$ modes on average. We observe that MB-OT and SB-CFM benefit most from \methodName{}. We hypothesize this is because MB-OT and SB-CFM are formulated with the notion of optimal paths, which results in a more accurate estimation of $\hat{x}_1$ and thus a better diversity gradient.

\subsection{Comparison To Particle Guidance}
\label{sec:comp-to-pg}

\begin{figure}[!ht]
    \centering
    \begin{subfigure}[t]{0.49\linewidth}
        \centering
        \includegraphics[width=\textwidth, height=\textwidth]{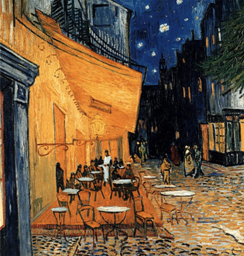}
        \caption{Training Data}
        \label{fig:pg-a}
    \end{subfigure}
    \begin{subfigure}[t]{0.49\linewidth}
        \centering
        \includegraphics[width=\textwidth, height=\textwidth]{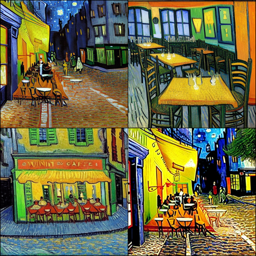}
        \caption{IID Sampling}
        \label{fig:pg-b}
    \end{subfigure}
    \begin{subfigure}[t]{0.49\linewidth}
        \centering
        \includegraphics[width=\textwidth, height=\textwidth]{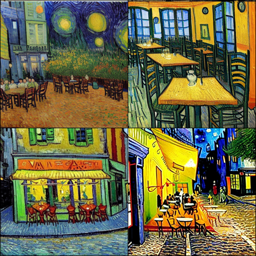}
        \caption{Particle Guidance \citep{corso2023particle}}
        \label{fig:pg-c}
    \end{subfigure}
    \begin{subfigure}[t]{0.49\linewidth}
        \centering
        \includegraphics[width=\textwidth, height=\textwidth]{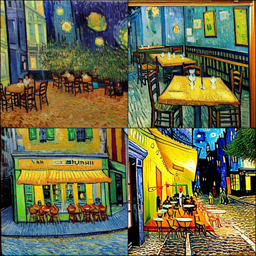}
        \caption{DiverseFlow}
        \label{fig:pg-d}
    \end{subfigure}
    \caption{For the prompt ``\texttt{VAN GOGH CAFE TERASSE copy.jpg}'', we show (a) the training data, (b) IID sampling with two copies in top-left and bottom-right, (c) Particle Guidance \citep{corso2023particle}, and (d) DiverseFlow.}
    \label{fig:memorize}
\end{figure}

\begin{figure}
    \centering
    \includegraphics[width=\linewidth]{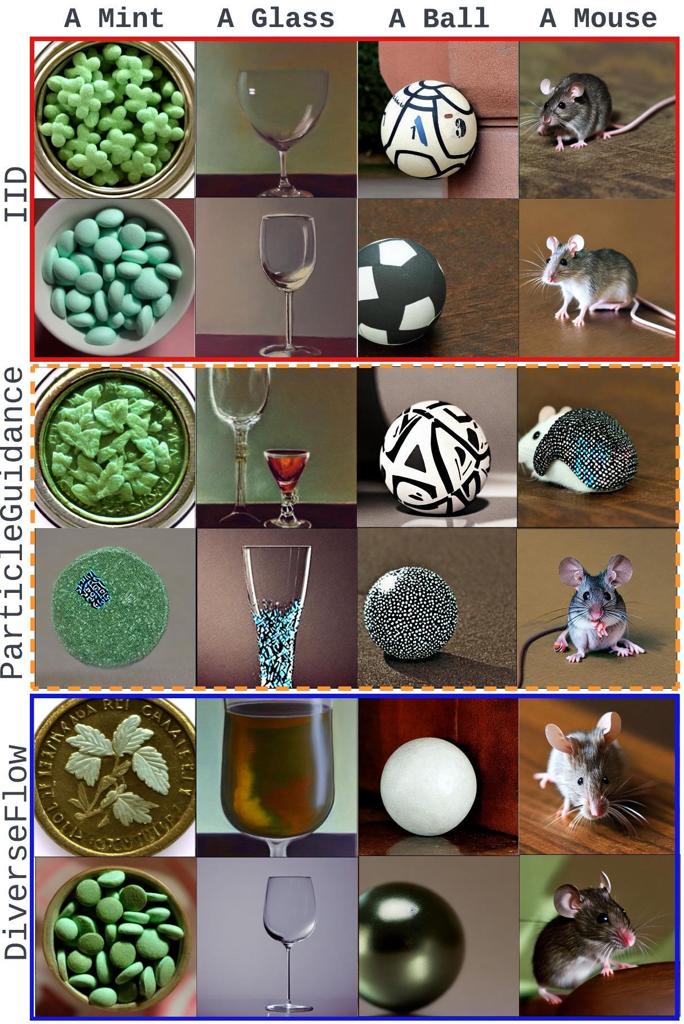}
    \caption{\footnotesize Particle Guidance (middle rows) can suffer from artefacts for polysemous prompts. \methodName{} retains quality while achieving diversity: for ``A mint'', a minted coin with a mint leaf was discovered.}
    \label{fig:artefact}
\end{figure}

\begin{figure}
    \centering
    \includegraphics[width=0.9\linewidth]{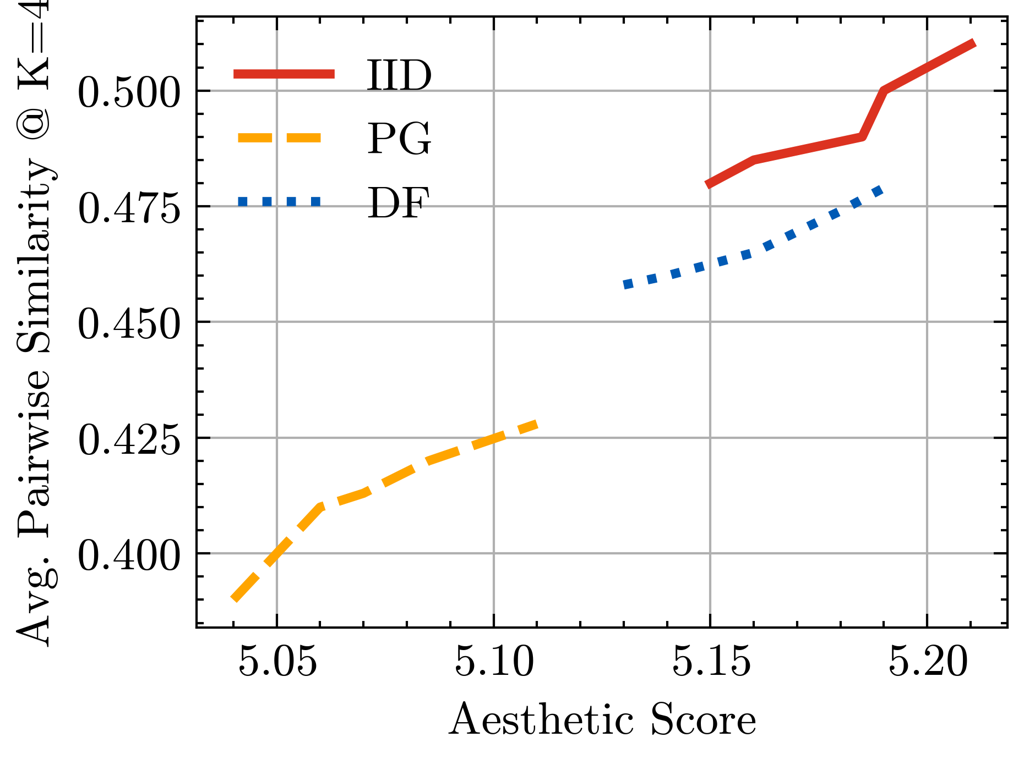}
    \caption{\footnotesize In-batch similarity $(\downarrow)$ vs aesthetic score $(\uparrow)$}
    \label{fig:pg-aes}
\end{figure}

\begin{figure}
    \centering
    \includegraphics[width=\linewidth]{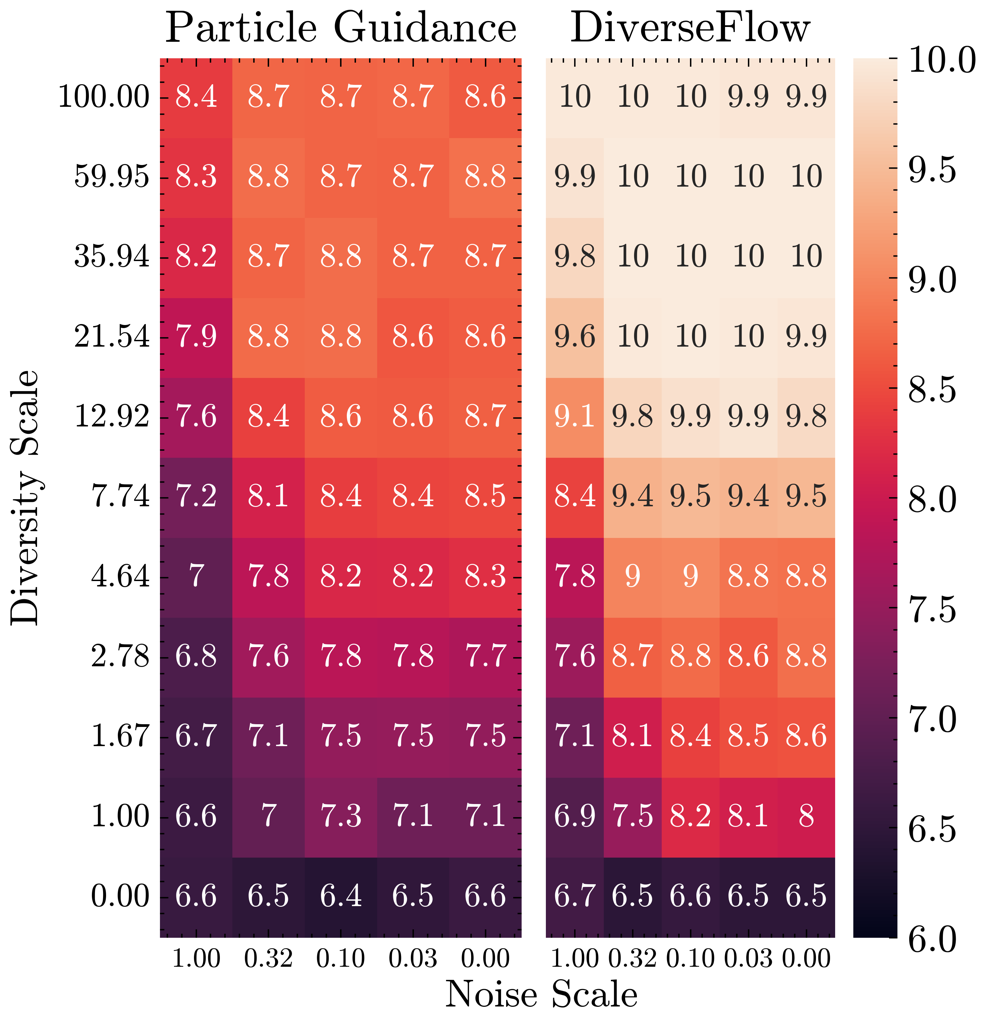}
    \caption{\footnotesize Average modes discovered over 100 trials, for varying noise scales and strengths of DiverseFlow and ParticleGuidance.}
    \label{fig:pg-heatmap}
\end{figure}

\noindent\textbf{(i) Preventing Training Data Copy:} Previously, \citet{corso2023particle} demonstrate that their method can alleviate Stable Diffusion's training data regurgitation problem \citep{somepalli2023diffusion} to some extent.
In \Cref{fig:memorize}, we demonstrate similar capabilities; in (b), the top left and bottom right examples are copies of the training data. Subsequently, (c) and (d) find a new example for the top-left sample. \textbf{(ii) Diversity vs Quality:} We also compare the diversity versus quality of \methodName{} against Particle Guidance in \Cref{fig:pg-aes} over 30 polysemous prompts repeated over 10 random seeds. 
While varying the strength of \methodName{} to improve diversity, we also vary the values of the classifier-free guidance strength from $7.5$ to $10$ to boost quality. Quality is measured by \emph{Aesthetic Score} (higher is better) \citep{aes}, and diversity is measured by \emph{average pairwise similarity} of a set (lower is better) \citep{corso2023particle}.
We observe that though \methodName{} obtains better diversity at similar quality to IID sampling, the aesthetic score of Particle Guidance is unusually low. In \Cref{fig:artefact} we highlight the cause: for the challenging task of polysemous text-to-image generation, the images generated from Particle Guidance (a) suffers from image artefacts, which achieves a high diversity score but a poor quality score. Additional details about the experiment, including the set of 30 prompts, are provided in the supplementary.
\textbf{(iii) Mode Discovery:} In high dimensional data, the number of modes, $N$, is significantly greater than the budget $K$, so finding unique modes is still highly probable. To better highlight the differences between \methodName{} and Particle Guidance, we adopt a simpler toy experiment: finding modes in a symmetric \emph{uniform} Mixture of Gaussian distribution. 
\citet{corso2023particle} provides the result that IID sampling with budget $K=10$ discovers \textbf{about 6.5 modes} on average, while Particle Guidance with a Euclidean kernel discovers \textbf{almost 9 out of 10 modes}. We verify this result in \Cref{fig:pg-heatmap}, finding that Particle Guidance discovers up to \textbf{8.8 modes} (averaged over 100 trials). However, by using \methodName{}, it is possible to discover \textbf{all 10 modes} on average, showing that our approach has a stronger diversification effect.

\section{Conclusion}
In this paper, we present \methodName{}, a way to enforce diversity among a limited set of samples generated from a flow by coupling them through a determinantal point process. We demonstrate multiple applications where our method can be useful. In essence, though flow models admit a deterministic mapping from source to sample, \methodName{} allows us to modify this mapping at inference-time and obtain samples with improved diversity. Diversity enhances the utility of the underlying generative flow in the case of ill-posed problems, where many solutions exist and it is desirable to find a multitude of such solutions. 

{
    \small
    \bibliographystyle{ieeenat_fullname}
    \bibliography{main.bib}
}

\clearpage
\setcounter{page}{1}
\maketitlesupplementary

In the supplementary, we primarily focus on two aspects: First, we present additional experimental details and ablation of the content in the main body of the paper. Second, we discuss some of the potential limitations of our method, as well as challenges and open questions.

\begin{figure*}
    \centering
    \includegraphics[width=\linewidth]{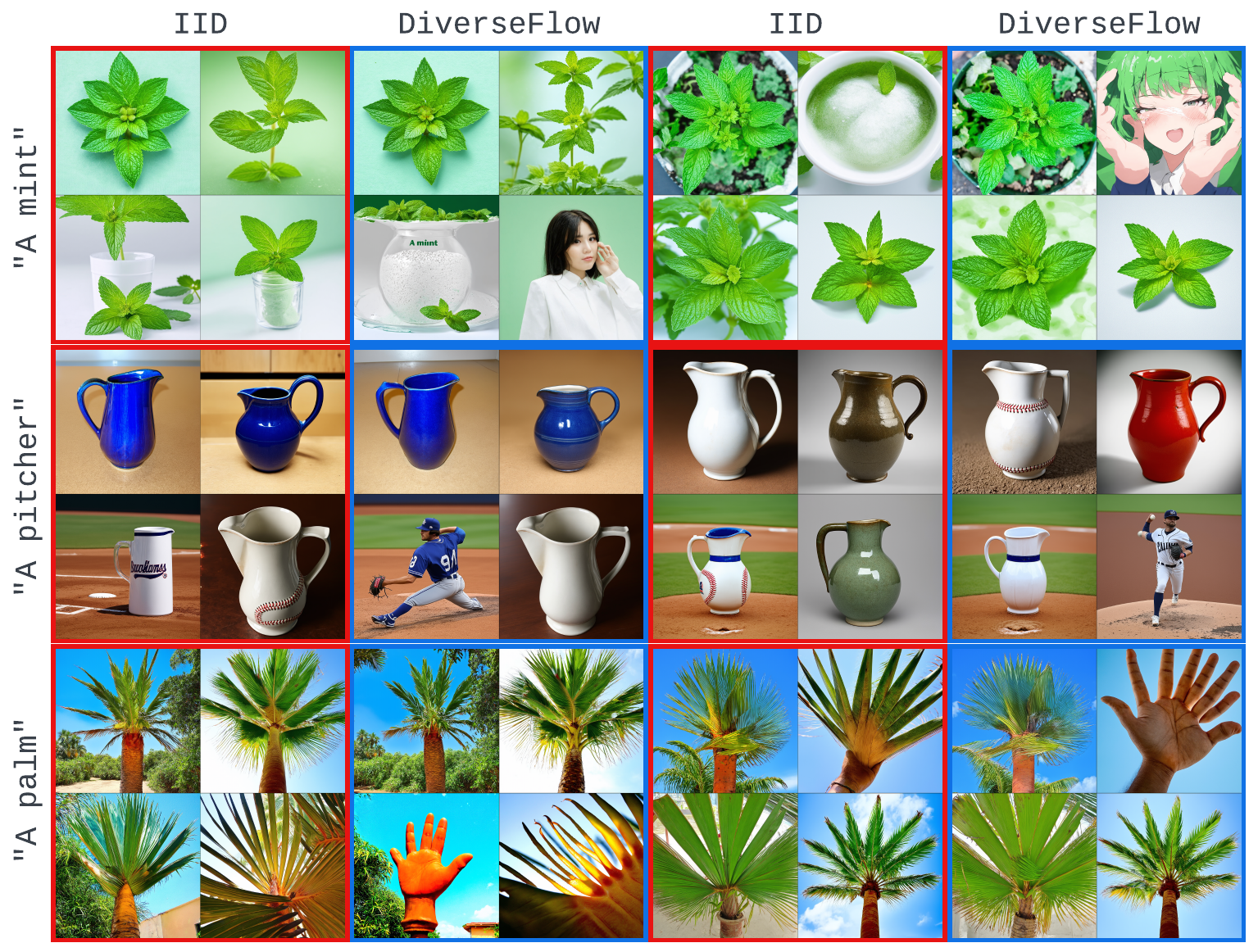}
    \caption{Some examples on SD3 where \methodName{} discovers alternate meanings that IID sampling doesn't find.}
    \label{fig:sd3-poly}
\end{figure*}

\section{Additional Experimental Details}
\label{appendix:sec:exp}

\subsection{Polysemous Prompts}
\label{appendix:sec:prompts}

\noindent\textbf{Rationale:} One question that may arise is why we use polysemous prompts to primarily evaluate the effects of \methodName{}, instead of some other existing regular text-to-image task. There are two reasons: (i) Diversity is clearly distinguishable both qualitatively and quantitatively for polysemous prompts (ii) Text-to-image generation from polysemous prompts is an inherently challenging task for generative flow ODEs.

\begin{figure}[ht]
    \centering
    \includegraphics[width=\linewidth]{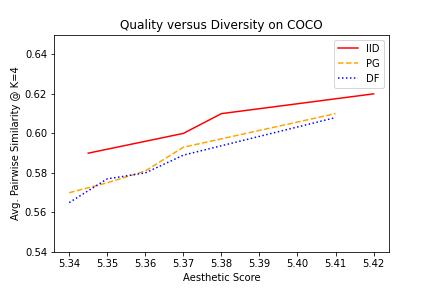}
    \caption{Diversity and Quality on COCO validation set.}
    \label{fig:coco-comp}
\end{figure}

\begin{figure}[ht]
    \centering
    \includegraphics[width=\linewidth]{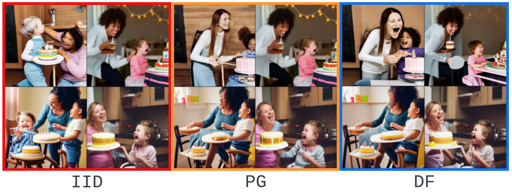}
    \caption{``A woman holding a cake is looking at an excited child in a high chair'': results from IID sampling, Particle Guidance, and \methodName{} respectively.}
    \label{fig:coco-example}
\end{figure}

In our early experiments, we considered the validation set from the COCO dataset as a way of evaluating diversity in text-to-image generation. However, although we observed an increase in diversity (average pairwise dissimilarity in a set), the difference was difficult to observe visually from images. For instance, in \Cref{fig:coco-example}, it is difficult to tell if diversity has improved from the original IID sample result. We also find in \Cref{fig:coco-comp} that \methodName{} and Particle Guidance achieve similar results, where it is difficult to distinguish between either.

In order to show an impactful example, we pose the task of text-to-image generation from ambiguous prompts that may carry multiple distinct meanings, with the assumption that multiple meanings would correspond to sufficiently disentangled modes in the data. In the case of polysemous prompts, the difference between diverse and non-diverse results is clear to the observer, and is significantly highlighted in the metrics. We also find that for the same guidance strength, Particle Guidance is highly prone to aliasing artifacts (as we show in \Cref{fig:artefact}); instead of finding diverse samples, it achieves higher dissimilarity by introducing noise in the image (hence the low similarity and low quality in \Cref{fig:pg-aes}). This suggests that open-ended prompts are inherently more difficult than well-defined and constrained prompts.

\noindent\textbf{Setup:} For direct comparison to Particle Guidance \citep{corso2023particle}, we utilize the probability flow ODE formulation of Stable Diffusion v1.5 \citep{ldm} as our underlying generative flow. We also apply \methodName{} on the larger Stable Diffusion v3 model \citep{esser2024scaling}, which is based on the rectified flow approach of \citet{liu2022flow}. We show some results for SD-v3 in \Cref{table:1}, and in \Cref{fig:sd3-normal} and \Cref{fig:sd3-poly}.

\newcommand{\xmark}{\ding{55}}

\begin{table*}[t]
\begin{center}
\begin{tabular}{ |c|c|c|c|c| } 
 \hline
 polysemous word & SD-v1.5 & SD-v1.5+DF & SD-v3 & SD-v3 + DF\\
 \hline
boxer  & \checkmark & \checkmark & \xmark & \checkmark \\
crane  & \checkmark & \checkmark & \checkmark & \checkmark\\
bat  & \xmark & \xmark & \xmark & \xmark\\
letter & \checkmark  & \checkmark & \checkmark & \checkmark \\
buck & \checkmark & \checkmark & \xmark & \xmark\\
seal & \checkmark & \checkmark & \xmark & \xmark  \\
mouse & \xmark & \xmark & \xmark & \xmark \\
horn & \checkmark  & \checkmark & \checkmark & \checkmark \\
chest & \xmark & \xmark & \xmark & \xmark \\
nail & \checkmark & \checkmark & \checkmark & \checkmark \\
ruler  & \xmark & \checkmark & \xmark & \checkmark\\
ball & \xmark  & \xmark & \xmark & \xmark\\
file & \checkmark  & \checkmark & \checkmark & \checkmark\\
ring & \xmark & \xmark & \xmark & \xmark \\
deck & \xmark  & \xmark & \xmark & \xmark\\
nut  & \xmark & \xmark & \xmark & \xmark\\
bolt & \xmark & \checkmark & \checkmark & \checkmark \\
bow & \xmark & \xmark & \xmark & \xmark \\
pupil & \xmark & \xmark & \xmark & \xmark \\
palm & \xmark & \checkmark & \checkmark & \checkmark\\
pitcher & \xmark & \xmark & \checkmark & \checkmark \\
fan & \xmark & \checkmark & \xmark & \xmark \\
club  & \checkmark  & \checkmark & \checkmark & \checkmark\\
anchor & \xmark & \xmark & \xmark & \xmark \\
mint & \checkmark & \checkmark & \xmark & \checkmark \\
iron & \xmark & \checkmark & \xmark & \checkmark\\
bank & \xmark  & \xmark & \xmark & \xmark\\
glass & \xmark & \xmark & \xmark & \xmark \\
pen & \xmark & \xmark & \xmark & \xmark \\
spring & \xmark & \xmark & \xmark & \xmark\\
\hline
total & 10 & 15 & 9 & 13 \\
\hline
\end{tabular}
\end{center}
\caption{List of polysemous prompts and possible discovered diverse meanings over 100 samples.}
\label{table:1}
\end{table*}

\noindent\textbf{Prompt Selection:} We adopt a set of 30 polysemous prompts, which are given in \Cref{table:1}. To find such prompts, we prompted an LLM for 50 polysemous nouns, and then we manually filtered 30 good polysemous words with clearly distinct meanings.

\noindent\textbf{Implementation:} We use 30 Euler steps to sample from SD-v1.5, and 28 Euler steps for SD-v3, with a classifier-free guidance strength of 8 and 7 respectively, which are the default settings of both models. For the feature extractor, we experiment with both CLIP-ViT-B16 and DINO-ViT-B8, and find better results with DINO. From \Cref{table:1}, it can be seen that polysemous prompts are a challenging task; for many prompts, it is not yet possible to find the diverse meanings. For example, for ``a spring'', both SD-v1.5 and SD-v3 only yield images of the season, and not the coiled object. \methodName{} helps discover 5 and 4 additional meanings for SD-v1.5 and SD-v3 respectively. For the images in \Cref{fig:artefact} and the results in \Cref{fig:pg-aes}, we use a scaling factor of $8\sigma(t)$ for Particle Guidance, same as used by the authors in their paper. For \methodName{}, we use $\frac{20\sigma(t)}{\norm{\nabla\log{\mathcal{L}(\m{x_t}^{(1)}, \m{x_t}^{(2)}, \dots, \m{x_t}^{(k)})}}}$.

\subsection{Inpainting}
\label{appendix:sec:inpaint}

\noindent\textbf{Rationale:} In masked face datasets, occlusion masks may occur in various areas of the face and in various sizes. In the case of small occlusion masks, or masks on insignificant ares (such as a cheek), there is a minimal scope for generating diverse results. We thus fix a large central mask that approximately covers 50\% of the face surface area, consisting primarily of the mouth and nose regions, as shown in \Cref{fig:inpaint}.

\begin{figure}[ht]
    \centering
    \includegraphics[width=\linewidth]{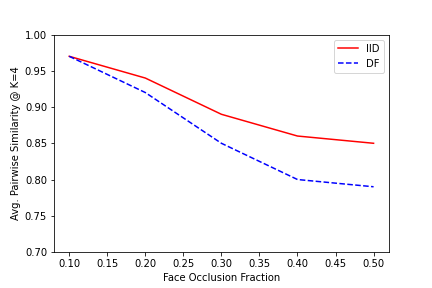}
    \caption{Similarity between faces decreases with increasing occlusion mask size. \methodName{} finds more dissimilar faces with larger occlusions, but has little effect on small occlusions.}
    \label{fig:scale-occ}
\end{figure}

\noindent\textbf{Setup:} We sample 500 random images from the CelebA-HQ $256\times 256$ dataset, and apply the same fixed mask to all images. Additionally, we vary how much of the face is occluded by the mask by scaling the size of the mask, to approximately cover $10\%$ to $50\%$ of the visible face. We then measure the average pairwise similarity between the generated faces ($K=4$ inpainting results per occluded face). In \Cref{fig:scale-occ}, we show that the effect of \methodName{} is limited for small occlusions, and is distinct for larger occlusions.

\noindent\textbf{Implementation:} To implement inpainting with an FM model, we utilize (i) an \emph{unconditional} off-the-shelf face image generating FM, and (ii) a continuous-time ODE inpainting algorithm. We adopt a RectifiedFlow model pre-trained on CelebAHQ-$256\times256$ \citep{karras2017progressive}, from \url{https://github.com/gnobitab/RectifiedFlow}. Next, we extend the manifold constrained gradient (MCG) algorithm \citep{chung2022improving} from diffusion models to FM models, in \Cref{algo:inpainting}. We use $\gamma(t) = 10\frac{\sqrt{1 - t}}{\norm{\nabla\log\mathcal{L}}}$ as a time-varying scale for the DPP gradient. Additionally, we use 200 Euler steps for sampling; more steps are needed in comparison to text-to-image generation for the MCG inpainting algorithm to converge.

For the feature encoder $F$, we use the FaRL model  \citep{zheng2022general}, which is a CLIP-like model trained on LAIONFace \cite{zheng2022general}. FaRL is trained in a mask-aware manner, and we downsample the inpainting mask to additionally create an attention mask, to ensure that the feature encoder $F$ does not focus on the irrelevant areas.

\begin{algorithm}
\caption{MCG Flow Inpainting with Euler Method}\label{alg:cap}
\begin{algorithmic}
\Require Inpainting input $\m{Y} \in \R^{H \times W \times 3}$, inpainting mask $\m{M} \in \mathbb{Z}_2^{H \times W \times 3}$, number of sampling steps $N$, time-varying velocity field $v_{\theta}$
\State $\m{X}_0 \sim \N(0, \I)$ \Comment{Sample a particle from source distribution $\m{Z}_0$}
\For{i=$0 \dots N-1$}
    \State $t_{i}, t_{i+1} \gets \frac{i}{N}, \frac{i + 1}{N}$ \Comment{Uniform timesteps, $t \in 0 \dots 1$}
    \State $\Delta_t \gets t_{i+1} - t_i$
    \State $\m{V}_{i} \gets v_\theta(\m{X}_{i}, t)$ \Comment{Predicted velocity at timestep $t$}
    \State $\hat{\m{X}}_N \gets \m{X}_i + \m{V}_{i} (1 - t)$ \Comment{Estimated target particle $\hat{\m{X}}_N \sim \m{Z}_1$}
    
    \State $\m{V}_{i} \gets \m{V}_i - \gamma(t) * \nabla_{\m{X}_i}\mathcal{LL}(\hat{\m{X}}_N)$ \Comment{DiverseFlow step}
    
    \State $\nabla_{\text{MCG}} \gets \frac{\partial}{\partial \m{X}_{i}}\norm{\m{Y} \odot \m{M} - \hat{\m{X}}_N \odot \m{M}}_2^2$ \Comment{Manifold Constrained Gradient}
    \State $\m{X}_{i + 1} \gets \m{X}_{i} + \m{V}_{i} \Delta_t$ \Comment{Euler step}
    \State $\m{X}'_{i + 1} \gets \m{X}_{i + 1} - \alpha_{t_i}\nabla_{\text{MCG}}$ \Comment{Apply MCG correction; $\alpha_{t_i} = \sqrt{1 - t_i}$}
    \State $\m{Y}_{i + 1} \gets \m{X}_0 (1 - t') + \m{Y} t'$ \Comment{Linearly interpolate between $\m{X}_0$ and $\m{Y}$ at $t_{i+1}$}
    \State $\m{X}''_{i + 1} \gets \m{X}'_{i + 1} \odot (1 - \m{M}) + \m{Y}_{i + 1}\odot \m{M}$ \Comment{Replace known region with $\m{Y}_{i + 1}$}
\EndFor
\State \Return $\m{X}_N$
\end{algorithmic}
\label{algo:inpainting}
\end{algorithm}

\subsection{Class-Conditioned Image Generation}
\label{appendix:sec:class-conditioned}

\begin{figure}
    \centering
    \includegraphics[width=\linewidth]{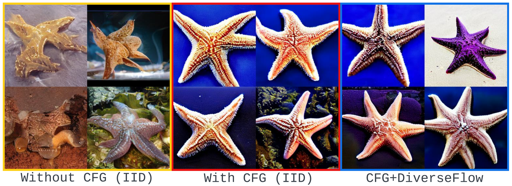}
    \caption{Generation for Class 327 (Starfish). DiverseFlow finds a significantly different result (a purple sea star) in top-right sample.}
    \label{fig:starfish}
\end{figure}

\textbf{Rationale:} Many ImageNet categories involve animal species that exhibit keen biodiversity. However, to observe the variation between species or animal families, we need to ensure diverse results. However, regular IID sampling can often be very strongly biased towards the dominant mode or variation (for instance, the scarlet Macaw in \Cref{fig:imagenet}, or the coral-shade starfish in \Cref{fig:starfish}). By improving the diversity of the generative model, we can easily discover more varieties with fewer number of samples.

\textbf{Setup:} We only show a few qualitative samples for class-label to image generation from ImageNet. In particular, we pick the classes `Macaw', `Mushroom', and `Starfish' as they are prominently demonstrated as examples in the project page of the underlying flow model (\url{https://vinairesearch.github.io/LFM}).

\textbf{Implementation:} For the ImageNet samples, we show in \Cref{fig:imagenet}, we use pre-trained LFM models \citep{dao2023flow}, specifically the `imnet\_f8\_ditb2' weights from \url{https://vinairesearch.github.io/LFM}. We primarily used DINO-ViT-B8 as the feature extractor $F$. 

\subsection{Mode Finding}
\label{appendix:sec:mode-finding}
We train a set of four identical models from scratch for the four FM variants used in \Cref{fig:modefinding}. Each model is an \emph{unconditional} generative model and is defined as an MLP consisting of 4 fully connected layers, each except the first having 256 hidden units; the first layer has a hidden size of 256 + 1 to account for the time input. We use the \texttt{torchcfm} library (\url{https://github.com/atong01/conditional-flow-matching}) for the conditional path construction.

We solve the ODE with an Euler solver with 100 steps. We start with a budget of $K=2$ (as for $K=1$, the ODE must always find at least 1 mode) and increase $K$ till $K = N = 10$, where $N=10$ is the true number of modes in the dataset. For each $K$, we repeat 1000 trials (by taking random seeds 0-999). We use $\gamma(t) = 2\frac{\sqrt{1 - t}}{\norm{\nabla\log\mathcal{L}}}$. Since the data is 2D, we do not use any feature encoder $F$.

We find $\sim 7$ modes on average with \methodName{}, while IID sampling finds $\sim 5.6$ modes. With regular IID sampling, the least diverse seems to be the Stochastic Interpolant \citep{albergo2023stochastic}. Additionally, for the quantity `maximum modes found at any trial' we observe that in over 1000 trials with a budget of $K=10$, IID sampling does not find a single instance of all 10 modes in any CFM formulation.

\subsection{Mode-finding With Ideal Score}
\label{appendix:sec:idealscore}
\begin{figure}[htpb]
    \centering
    \includegraphics[width=\linewidth]{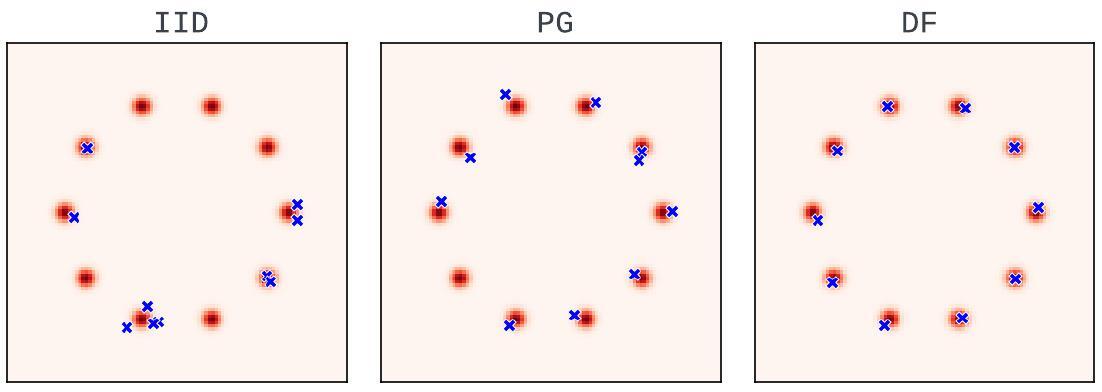}
    \caption{Finding modes on uniform mixture of Gaussian with true score.}
    \label{fig:truescore}
\end{figure}

In \Cref{fig:pg-heatmap}, no model is trained, and we have access to a true score function of a mixture of uniform Gaussian distribution, as shown in \Cref{fig:truescore}. We scale the DPP gradient by $\gamma(t) = W\frac{\sigma(t)}{\norm{\nabla\log\mathcal{L}}}$, where $\sigma(t)$ is the variance schedule path, and $W$ is a variable temperature parameter (Diversity Scale or Y-axis in \Cref{fig:pg-heatmap}). We also vary the noise levels from 1 (regular SDE) to 0 (probability flow ODE); it can be observed in \Cref{fig:pg-heatmap} that both Particle Guidance and DiverseFlow find the best result at noise level of 0.1.

\subsection{Choice of Feature Extractor}

\begin{figure}[ht]
    \centering
    \includegraphics[width=0.8\linewidth]{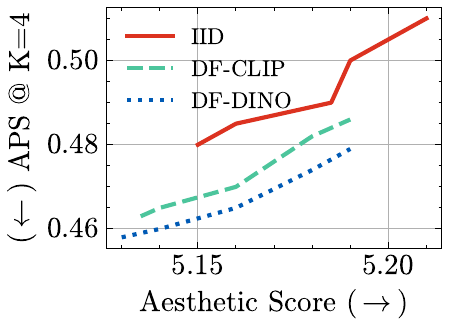}
    \caption{CLIP versus DINO feature extractor.}
    \label{fig:abl-feat}
\end{figure}

\Cref{fig:abl-feat} shows an ablation over the effect of using a CLIP vs. a DINO feature extractor. We observe that DINO achieves better diversity and quality on the polysemous prompts (\Cref{table:1}). This may be due to the fact that using DINO results in a purely image-based feature similarity. However, CLIP features are trained with image-text similarity, and might struggle with polysemous images. For example, an image of a human boxer and an image of a boxer dog can bothmap to similar CLIP latents, despite having stark visual differences.

\begin{figure*}[ht]
    \centering
    \includegraphics[width=0.8\linewidth]{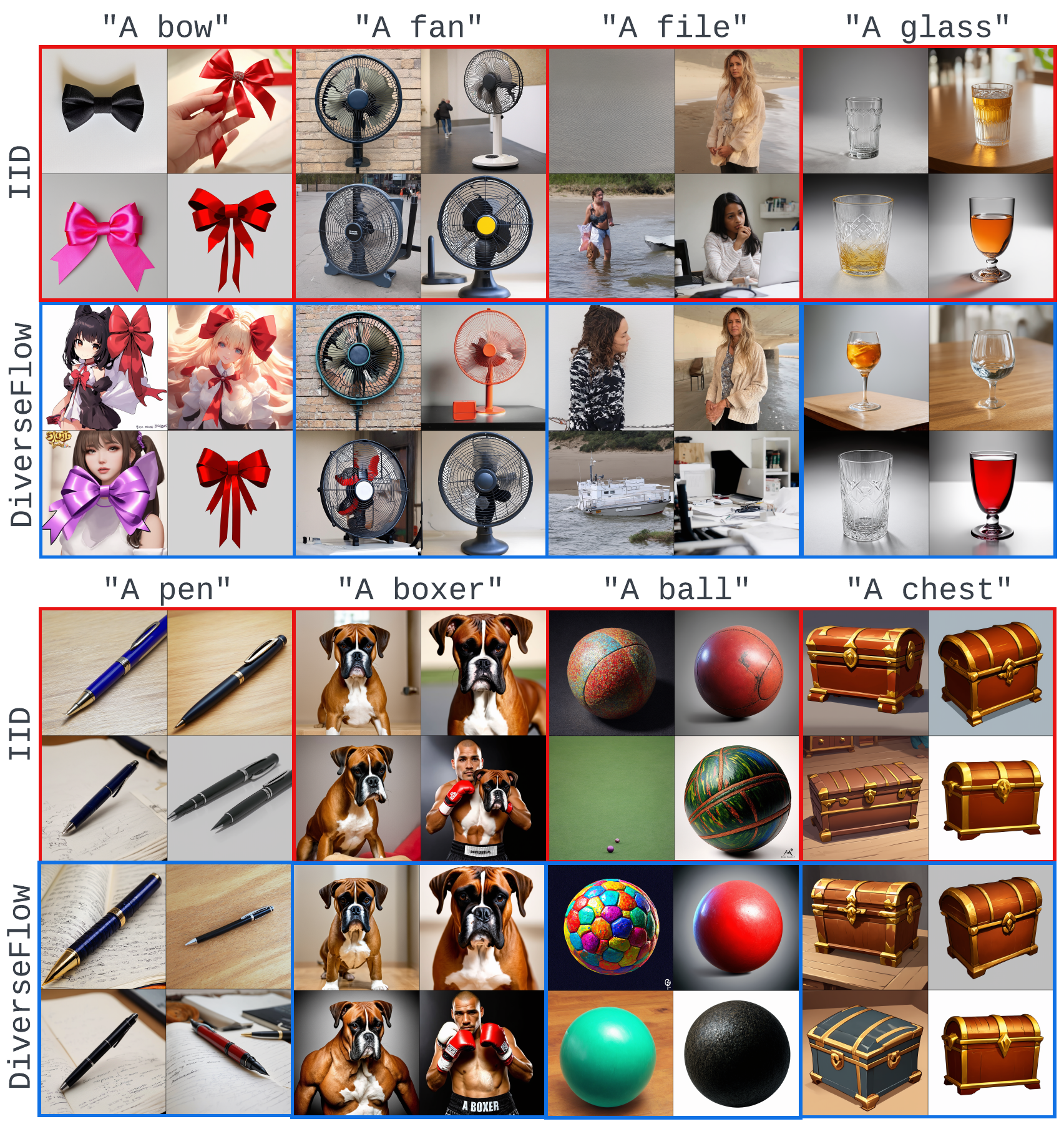}
    \caption{Some examples on SD3 where significantly polysemous meanings are not discovered. However, DiverseFlow still yields more diverse samples compared to IID samples.}
    \label{fig:sd3-normal}
\end{figure*}

\subsection{Connections to Particle Guidance}
\label{app:compare-to-pg}

It is possible to formulate Particle Guidance in DiverseFlow's framework. Consider the DPP kernel $\m{L}$ that we define in \Cref{eqn:dpp-kernel}. Particle Guidance defines a time-varying `log potential' that takes the form:

\begin{equation}
    \log\Phi_t^{(i)}(\m{x}^{(1)}, \m{x}^{(2)}, \dots, \m{x}^{(k)}) = \sum_j \m{L}^{(ij)}
\end{equation}

\begin{figure}[ht]
    \centering
    \begin{subfigure}[t]{0.4\linewidth}
        \centering
        \includegraphics[width=0.8\textwidth, height=0.8\textwidth]{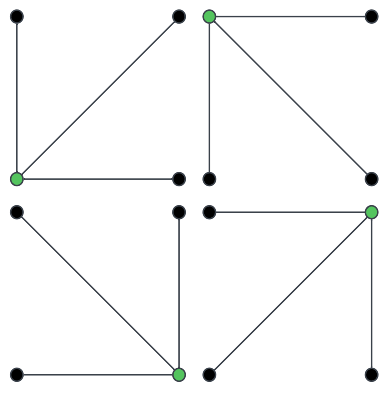}
        \caption{Computing the sum of pairwise distances from current particle (green) to every other particle (black)}
        \label{fig:pg}
    \end{subfigure}\quad
    \begin{subfigure}[t]{0.4\linewidth}
        \centering
        \includegraphics[width=0.8\textwidth, height=0.8\textwidth]{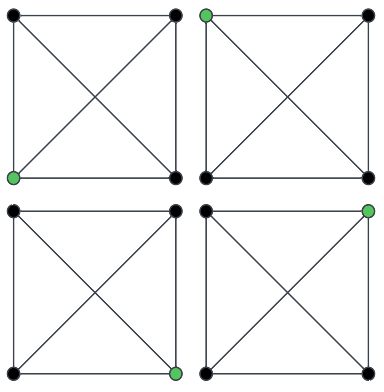}
        \caption{Computing the determinant needs to consider the distance of every point from every other point}
        \label{subfig:dpp}
    \end{subfigure}
    
    \caption{A geometric look at Particle Guidance and \methodName{}.}
    
    \label{fig:pgvsdpp}
\end{figure}

That is, the log potential for each particle is its pairwise similarity with every other particle. However, it is not readily apparent why the log potential is this pairwise sum (Equation 4 in particle guidance paper). In our work, the DPP is a probability measure that yields an approximate likelihood of the joint distribution $p(\m{x}^{(1)}, \m{x}^{(2)}, \dots, \m{x}^{(k)})$. Therefore, the log potential is simply the log-likelihood of the DPP. One geometric way to interpret the two approaches may be observed in \Cref{fig:pgvsdpp}.

Thus, the log potential for each particle in particle guidance is distinct. However in our work, the potential is the same for any particle, as it is defined over the determinant. The kernel-sum utilized in Particle Guidance can also be interpreted as an approximate joint likelihood function, except, unlike the DPP, it assigns a non-zero likelihood to the occurrence of duplicate elements. It is thus a softer form of diversification, which can be observed in \Cref{fig:pg-heatmap}. Finally, particle guidance does not consider a quality factor on the kernel, unlike DPP-based methods.

\subsection{Connections to training-based approaches}
There are several training-based approaches that implicitly improve diversity. For instance, assigning more optimal coupling \citep{tong2023improving, li2024immiscible} can reduce the distance between data and noise, which makes it unlikely for different source samples to be coupled with the same target---thereby improving both the quality and diversity in expectation.

One may question whether it is possible to directly train coupled ODEs, such as the one defined in \Cref{eqn:guidance}. To do so, it is necessary to make modifications in the model architecture, such that each individual point becomes aware of other points in the set/batch. Video-based generative models introduce a similar type of coupling between frames by adding temporal attention, and can be used as inspiration. In essence, converting \methodName{} to a trainable approach would require learning a time-varying $K \times d$ matrix field, which is a relatively unexplored area in generative learning; relevant research that explores this direction is the recent work of \citet{isobe2024extended}, that extends flow matching over matrix fields. We hope to explore training-based approaches that incorporate determinantal point processes in future work.

\section{Limitations and Challenges}
\label{appendix:sec:limitations}

\subsection{Soft-DPP Objective:}
 Recall that the DPP assigns a zero likelihood to a set $\{x^{(1)}, x^{(2)}, \dots, x^{(k)}\}$ as long as any $x^{(i)} = x^{(j)}$, that is, duplicate elements are not tolerated.

The exact log-likelihood defined in \Cref{eqn:dpplikelihood} can be thus be undefined on the rare occasion when we have near-identical elements in the set. The work of \citet{yuan2019diverse} presents a relaxed objective to address this problem. Instead of maximizing $\sum_a\log(\lambda_a / (1 + \lambda_a))$, we can maximize the expectation of the cardinality of the DPP (a bound on the rank of $\m{L}$):

\begin{equation}
\begin{split}
    \E\left[\mid \{\hat{\m{x}}_1^{(1)}, \hat{\m{x}}_1^{(2)}, \dots, \hat{\m{x}}_1^{(k)}\} \mid\right] &= \sum_{a=1}^k \frac{\lambda(\m{L})_a}{\lambda(\m{L})_a + 1}\\
    &= \text{Tr}(\I - (\m{L} + \I)^{-1} )
    \label{eqn:dpp-approx}
\end{split}
\end{equation}

For high-dimensional problems (such as text-to-image generation), we find that the exact likelihood \Cref{eqn:dpplikelihood} is suitable, as it is highly unlikely for random source points to be identical in high-dimensional space. 

\subsection{Limited by Underlying Model}
From a modeling perspective, while \methodName{} seeks to enhance the sample diversity of flow-matching models under a fixed sampling budget, it is still limited by the distribution modes the underlying FM models have learned.
For instance, the word ``mouse" may refer to: (i) a mammal (rodent), (ii) a computer peripheral. \methodName{} could not generate any samples of the computer mouse with just the prompt ``a mouse'' ( \Cref{fig:artefact}); we hypothesize that the learned likelihood of the animal significantly dominates the latter meaning. Again, with SD-v3, we could not find any examples of coins for ``a mint'' which we could find for SD-v1.5. Thus, the discovery of diverse modes is still clearly dependent on the model being used. In \Cref{fig:sd3-normal}, we show some examples where the polysemous meaning was not discovered, and in \Cref{fig:sd3-poly}, we show discovered polysemous meanings.

\begin{algorithm}[ht]
\caption{Progressively Growing Kernel}\label{alg:prog}
\begin{algorithmic}
\Require number of progressions $R$, number of sampling steps $N$, time-varying velocity field $v_{\theta}$, budget $K$
\State $C = \{\}$ \Comment{Initialize Cache}
\For{r=$0 \dots R-1$}
\State $\m{X}_0 \sim \N(0, \I)$ \Comment{Sample source}
\State $S \gets \mid C \mid$ \Comment{Cache size}
\For{i=$0 \dots N-1$}
    \State $t_{i}, t_{i+1} \gets \frac{i}{N}, \frac{i + 1}{N}$ \Comment{Uniform timesteps}
    \State $\Delta_t \gets t_{i+1} - t_i$
    \State $\m{V}_{i} \gets v_\theta(\m{X}_{i}, t)$ \Comment{velocity at timestep $t$}
    \State $\hat{\m{X}}_N \gets \m{X}_i + \m{V}_{i} (1 - t)$ \Comment{Estimated target}
    \State $\m{X} \gets \{\hat{\m{X}}_N^{(1)}, \dots, \hat{\m{X}}_N^{(K)}, C^{(1)}, \dots, C^{(S)}  \}$ \Comment{Add cached samples to the set}
    \State $\m{V}_{i} \gets \m{V}_i + \gamma(t) * \nabla_{\m{X}_i}\mathcal{LL}(\m{X})$ \Comment{DiverseFlow step}
    
    \State $\m{X}_{i + 1} \gets \m{X}_{i} + \m{V}_{i} \Delta_t$ \Comment{Euler step}
\EndFor
\State $C \gets \m{X}_N$ \Comment{Add to cache}
\EndFor
\State \Return C
\end{algorithmic}
\end{algorithm}

\subsection{Computational Cost}
From a computational perspective, for high-resolution generative modeling, estimating the diversity gradient $\nabla_{\m{x}_t}\mathcal{LL}$ can be memory intensive. With either Stable Diffusion or LFM, it is necessary to backpropagate over (i) the KL-regularized AutoEncoder, (ii) the feature encoding ViT, $F$, and (iii) the high-resolution sample $\hat{\m{x}_1}$---thus practically limiting us to a batch of 4 samples at a time. We note that Particle Guidance faces a similar challenge.

One way to overcome the memory limit is to utilize a progressively growing kernel: we can sample a set of 4 images, and then sample another 4, where the kernel is $8\times8$, and another 4, where the kernel is $12 \times 12$, and so on. Thus, only the kernel size will grow to $K \times r$ at any iteration $r$, but we will still compute the gradient with respect to $K$ samples. We provide a pseudocode for this procedure in \Cref{alg:prog}.

\subsection{Entangled Modes}
We find that in many cases, the diverse results obtained by \methodName{} consist of multiple semantic meanings entangled into one image (for instance, coin with deer head, or or coin with mint leaves). However, we find that these entangled modes are a characteristic of the generative models for polysemous prompts, and thus also a limitation of the underlying model.

An open question for the research community can be how to induce diversity so that disentangled and distinct modes are discovered for ambiguous prompts, rather than entangled ones. Further, numerically measuring the entanglement of different concepts in a generated image could be an initial step towards solving this problem.

\end{document}